\documentclass[11pt]{article}

\usepackage{tcolorbox}
\usepackage{enumitem}

\usepackage{amsmath}
\usepackage{amssymb}
\usepackage[table]{xcolor}
\usepackage{tikz}

\usepackage{pgfplots}
\pgfplotsset{compat=1.17}

\usepackage{hyperref}

\usepackage{booktabs}
\usepackage{tabularx}
\usepackage{array}
\usepackage{seqsplit}
\usepackage{makecell}
\usepackage{longtable}
\usepackage{xurl}

\usepackage{float}

\usepackage{algorithm}
\usepackage{algpseudocode}

\definecolor{methodblue}{RGB}{232,244,255}
\newcommand{\best}[1]{\textbf{#1}}
\newcommand{\second}[1]{\underline{#1}}
\newcommand{\oursrow}{\rowcolor{methodblue}}
\usepackage{acl}

\usepackage{times}
\usepackage{latexsym}

\usepackage[T1]{fontenc}

\usepackage[utf8]{inputenc}

\usepackage{microtype}

\usepackage{inconsolata}

\usepackage{graphicx}
\graphicspath{{../}}

\title{What Should a Skill Remember? Quality-Cost Trade-offs in Cost-Aware Skill Rewriting for Language Model Agents}

\author{
\begin{tabular}{c}
\textbf{
Qinghua Xing\textsuperscript{1}\thanks{This work was done during Qinghua Xing's internship at Huawei Technologies.} \quad
Yinda Chen\textsuperscript{1} \quad
Yaping Jin\textsuperscript{2} \quad
Zhenhe Wu\textsuperscript{2} \quad
Bohan Lin\textsuperscript{1}
} \\
\textbf{
Hang Zhou\textsuperscript{3} \quad
Xinghao Chen\textsuperscript{2} \quad
Hanting Chen\textsuperscript{2}\thanks{Corresponding author: \texttt{chenhanting@huawei.com}.} \quad
Zhiwei Xiong\textsuperscript{1}
} \\
{\normalfont \textsuperscript{1}University of Science and Technology of China}\\ 
{\normalfont \textsuperscript{2}Huawei Technologies} \quad 
{\normalfont \textsuperscript{3}Tianjin University}
\end{tabular}
}

\begin{document}
\maketitle
\begin{abstract}
Large language model agents increasingly rely on skills: reusable procedural documents encoding workflows, tool use, implementation patterns, validation checks, and domain rules. Skill rewriting is often treated as prompt compression, but shorter skills can make agents more expensive by removing sparse operational anchors that prevent exploration, debugging, and recovery. We study skill rewriting through this economic lens. Our controlled framework profiles skill structure, rewrites skills using information-preservation strategies, and evaluates the rewrites under fixed task instructions, environments, and verifiers. Experiments on SkillsBench reveal distinct quality--cost trade-offs across strategies: API/code anchoring, workflow guarding, and rule/formula anchoring benefit different task families, with no universally dominant template. In the main held-out evaluation, the learned policy reduces total cost by 7.0\% and downstream agent-token cost by 6.0\%; in frozen cross-model transfer, the corresponding reductions average 14.7\% and 13.7\%, while verifier quality is preserved. These results position skill design as cost-aware operational knowledge engineering rather than prompt compression. Resources: \href{https://github.com/1Reminding/Skill_EE}{SkillEE}.

\end{abstract}

\section{Introduction}

Large language model (LLM) agents increasingly solve tasks by combining model reasoning with external tools, code execution, web interfaces, and environment feedback \citep{yao2023react,schick2023toolformer,qin2024toolllm,yang2024sweagent}. In these systems, performance depends not only on the base model, but also on the operational context, interfaces, and procedural knowledge supplied to the agent \citep{liu2024agentbench,zhou2024webarena,jimenez2024swebench,yao2024taubench}. A growing abstraction for providing such knowledge is the \emph{skill}: a reusable procedural document or bundle that encodes workflows, tool usage, code patterns, examples, validation checks, and domain rules \citep{wang2023voyager,anthropic2025agentskills,openai2026skills,li2026skillsbench}. Skills can substantially affect agent behavior, yet we still lack a systematic understanding of how their written structure should be revised when the goal is to reduce cost without degrading execution.
\begin{figure}[t]
    \includegraphics[width=\linewidth]{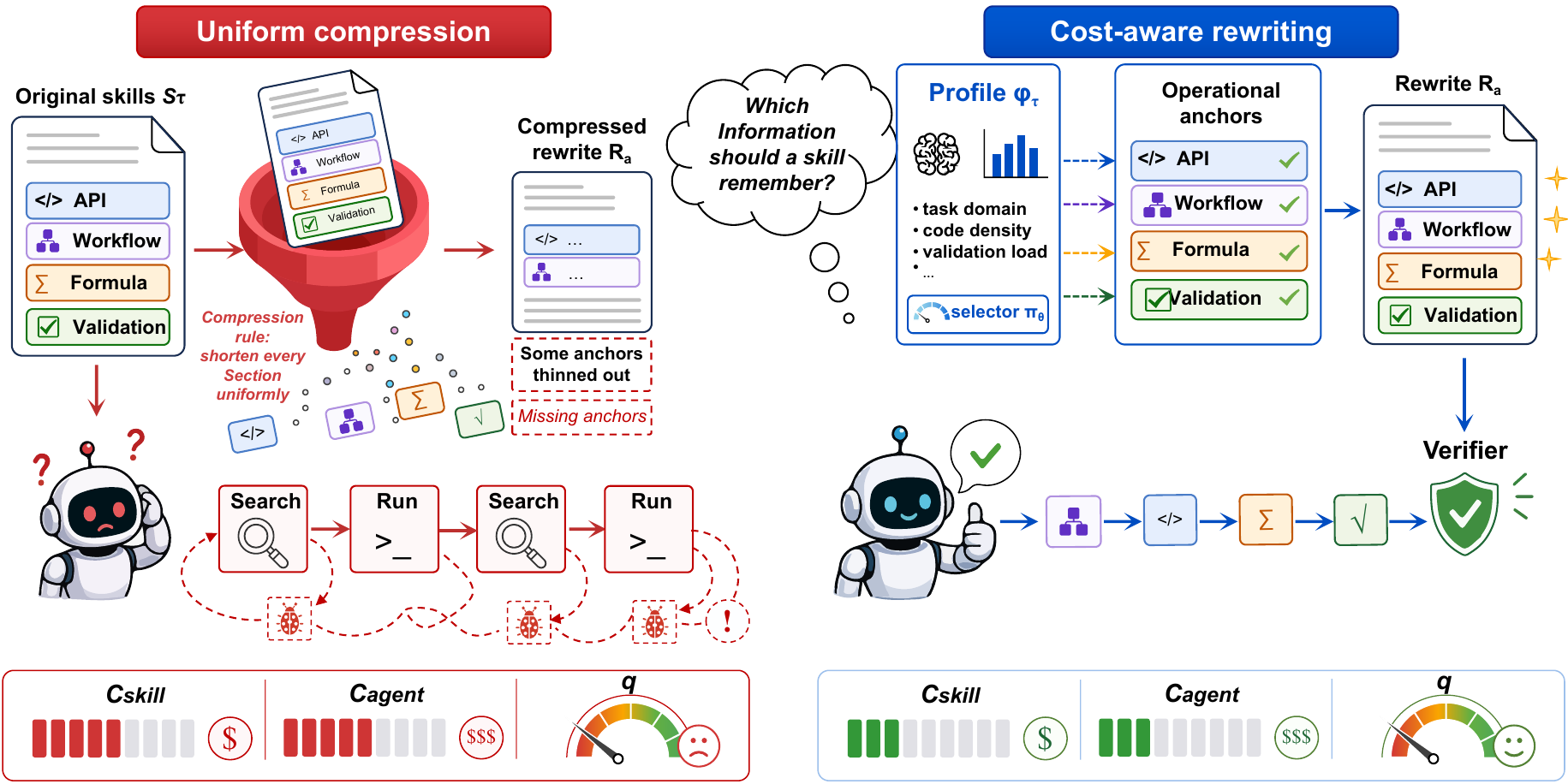}
    \caption{
Skill rewriting as cost-aware preservation rather than text compression.
Uniform compression may remove operational anchors and increase downstream execution cost, whereas policy-selected rewriting retains task-relevant anchors to better balance quality and total execution cost.
}
    \label{fig:intro-motivation}
\end{figure}
A natural approach is to shorten skills, especially because long prompts, retrieved contexts, and interaction histories increase inference cost \citep{li2023selectivecontext,jiang2023llmlingua,jiang2024longllmlingua,li2025promptcompressionsurvey,xiao2026agentdiet}. Prompt-optimization systems further show that instructions and LLM pipelines can often be improved through search, feedback, or compilation \citep{pryzant2023automatic,fernando2023promptbreeder,khattab2024dspy}. However, skills are not ordinary prompts or retrieved contexts. Their usefulness often depends on sparse \emph{operational anchors}: an API constructor, a command-line flag, a validation threshold, a file-format convention, a formula, or a recovery rule. Removing surrounding explanation may reduce direct skill-token cost, but removing the wrong anchor can increase downstream exploration, debugging, tool calls, or recovery. In our held-out evaluation, for example, a fixed workflow-oriented rewrite shortens skill documents but raises downstream agent-token usage to $1.14\times$ the original-skill baseline. Thus, skill rewriting poses an economic question: \emph{which information should a skill remember?}

We argue that the answer is task-dependent. Implementation-heavy skills may require API signatures, imports, object construction patterns, and code snippets; validation-heavy workflows may require ordered checks and failure-recovery cues; rule-governed tasks may require formulas, thresholds, schemas, and tie-breaking conventions. Prior skill benchmarks and analyses show that skill utility varies across domains, task structures, and models \citep{li2026skillsbench,liu2026skillswild,zhong2026skilllearnbench}. These differences imply that a single fixed rewrite template is unlikely to be uniformly optimal. Instead, skill rewriting should identify which operational anchors are predictive of quality and execution cost for a given task, and preserve those anchors accordingly.

This trade-off requires measuring rewriting at the level of agent execution, not only at the level of input length. Cost-aware LLM systems reduce expense through prompt adaptation, model cascades, and routing \citep{chen2023frugalgpt,yue2024large,ong2024routellm,shirkavand2025costaware}, while agent-cost work highlights that execution trajectories themselves can dominate token usage \citep{xiao2026agentdiet}. Skill rewriting sits between these regimes: removing text saves direct skill tokens, but removing the wrong details may shift cost into API exploration, code repair, tool reruns, or verifier recovery. We therefore view skill rewriting as cost-aware knowledge preservation, where the objective is to preserve the procedural details that matter for both quality and total execution cost.

We study this problem in SkillsBench \citep{li2026skillsbench} by rewriting only the skill documents while keeping task instructions, environments, and verifiers fixed. We compare preservation strategies that emphasize API/code details, workflow guards, and rules or formulas, and measure task quality together with direct skill-token, downstream agent-token, and total cost. Experiments show that no single rewrite structure is uniformly best; instead, different task and skill profiles favor different preserved anchors. On the primary held-out panel, a lightweight task-conditioned policy selects among these strategies and maintains verifier quality while reducing both direct skill-token and downstream agent-token cost. In the main held-out evaluation, the learned policy reduces total execution cost by 7.0\% and downstream agent-token cost by 6.0\%. When the same frozen policy is transferred across additional agent stacks, these reductions average 14.7\% and 13.7\%, respectively, while verifier quality is maintained or slightly improved. Our main contributions are:

\begin{enumerate}
    \item We formulate skill rewriting as a cost-aware preservation problem and introduce a controlled framework for profiling, rewriting, auditing, and evaluating agent skills under fixed task instructions, environments, and verifiers.
    \item We define economic metrics for agent skills that separate task quality, direct skill-token cost, downstream agent-token cost, total execution cost, and execution overrun, enabling analysis beyond skill length alone.
    \item We show that effective rewrite strategies depend on task and skill structure, and learn a task--skill-conditioned selector that chooses preservation strategies from structural features to improve the quality--cost trade-off.
\end{enumerate}

\section{Related Work}
\paragraph{Agents, tools, and procedural skills.}
LLM agents extend language models from single-turn prediction to interactive execution, where models reason, call tools, execute code, browse web interfaces, or operate in software-engineering environments. Early work such as ReAct formalized the reasoning--action loop \citep{yao2023react}, while Toolformer, API-Bank, and ToolLLM study tool/API use as a learnable and evaluable capability \citep{schick2023toolformer,li2023apibank,qin2024toolllm}. This shift has led to benchmarks and systems that evaluate agents through environment interaction, including multi-domain agent tasks \citep{liu2024agentbench}, web navigation \citep{zhou2024webarena}, software repair and agent-computer interfaces \citep{jimenez2024swebench,yang2024sweagent}, and tool-agent-user transactions \citep{yao2024taubench}. Beyond tool access itself, recent work increasingly externalizes reusable procedural knowledge as \emph{skills}. Voyager stores executable code skills for embodied exploration \citep{wang2023voyager}, and recent agent platforms package skills as bundles of instructions, scripts, examples, and resources \citep{anthropic2025agentskills,agentskills2026standard,openai2026skills}. SkillsBench further formalizes Agent Skills as structured procedural documents evaluated with deterministic verifiers across diverse tasks \citep{li2026skillsbench}, while recent surveys frame skill acquisition, retrieval, composition, and governance as central problems for skill-centric agents \citep{xu2026agentskills,jiang2026agenticskills,zhou2026agentskillssurvey}. These studies establish skills as an important interface for procedural knowledge, but they largely treat skill content as given or evaluate skills after creation. Our work instead studies the internal writing structure of skills: which operational anchors should be preserved during rewriting, and how those choices affect task quality, downstream execution, and total cost.

\paragraph{Prompt compression and prompt optimization.}
Prompt compression reduces LLM inference cost by removing or re-encoding less salient input content. Early methods prune low-information context \citep{li2023selectivecontext}, while LLMLingua, LongLLMLingua, and LLMLingua-2 estimate token importance or learn task-agnostic compression objectives for prompts and long contexts \citep{jiang2023llmlingua,jiang2024longllmlingua,pan2024llmlingua2}. Other work compresses contexts into learned memory representations \citep{ge2024icae}. In parallel, prompt optimization methods search, edit, or compile prompts and LLM pipelines using textual feedback, evolutionary search, or task-level supervision \citep{pryzant2023automatic,fernando2023promptbreeder,khattab2024dspy}. These approaches mainly optimize input length, semantic retention, or task accuracy for prompts and contexts, leaving less explored how compression affects procedural documents whose value may reside in sparse but execution-critical details.

\paragraph{Cost-aware LLM and agent systems.}
Cost-aware LLM systems optimize quality--cost trade-offs through prompt adaptation, model cascades, and routing \citep{chen2023frugalgpt,yue2024large,ong2024routellm,shirkavand2025costaware}. Prompt-compression surveys further distinguish hard token-level compression from soft representation-based compression \citep{li2025promptcompressionsurvey}. In agent settings, however, cost is also trajectory-dependent. Interface design has been shown to affect agents' ability to navigate repositories, edit files, and execute tests \citep{yang2024sweagent}. Rewriting a skill can reduce direct skill-token cost while increasing downstream execution cost. Structured skills can substantially change agent performance across tasks \citep{li2026skillsbench}; and recent trajectory-reduction work identifies growing interaction histories as a major source of input-token cost \citep{xiao2026agentdiet}. These findings broaden cost-aware design from model routing to the operational artifacts that shape agent execution.

\section{Skill Rewriting as Cost-Aware Knowledge Engineering}
\label{sec:method}
\subsection{Overview and Notation}
\begin{figure*}[t]
    \centering
    \includegraphics[width=\linewidth]{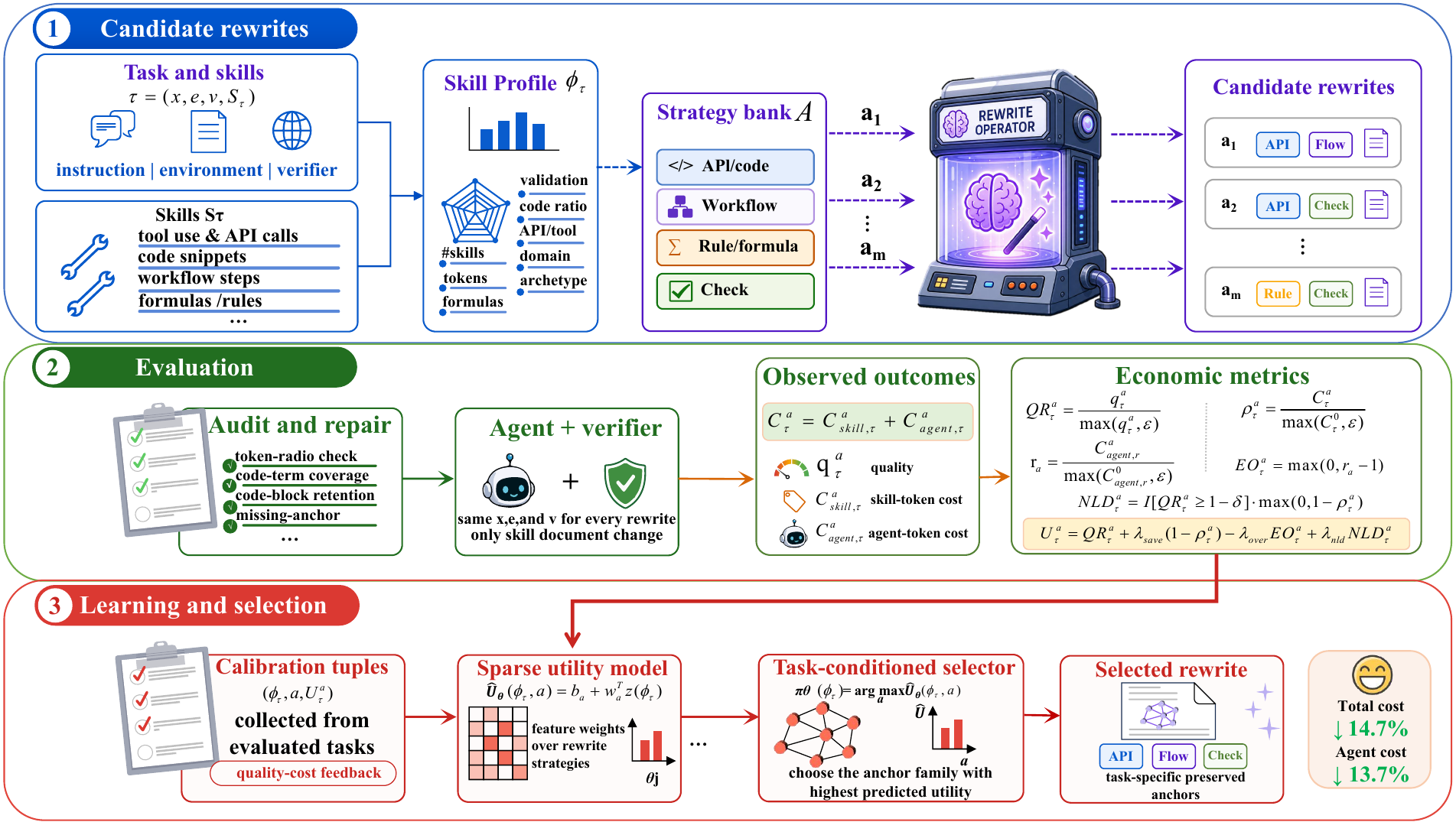}
    \caption{
    Overview of cost-aware skill rewriting.
    The pipeline profiles task skills, generates strategy-conditioned rewrites, evaluates them under fixed task conditions, and learns a task-conditioned strategy selector from quality--cost feedback.
    Arrows denote rewritten-skill flow, evaluation signals, and policy feedback.
    }
    \label{fig:framework}
\end{figure*}

We formulate skill rewriting as a cost-aware preservation problem. An agent task is $\tau=(x,e,v,\mathcal{S}_{\tau})$, where $x$ is the instruction, $e$ is the execution environment, $v$ is the verifier, and $\mathcal{S}_{\tau}=\{s_1,\ldots,s_m\}$ is the skill set. A rewrite strategy $a\in\mathcal{A}$ produces rewritten skills $\widetilde{\mathcal{S}}_{\tau}^{a}=R_a(\mathcal{S}_{\tau})$. Throughout rewriting and evaluation, the task instruction, environment, and verifier are fixed; only the skill documents are changed. This controlled setup isolates skill wording and structure as the experimental variable.

Running an agent with $\widetilde{\mathcal{S}}_{\tau}^{a}$ produces a verifier outcome and token-usage logs. These outcomes are converted into the economic utility used for strategy learning, with the metric definitions given in Section~\ref{sec:experimental-setup}. The method below specifies how skills are profiled, rewritten, audited, and selected by a task-conditioned policy.

\subsection{Skill Profiling and Rewrite Strategies}
We compute a structural profile $\phi_{\tau}=\phi(\tau,\mathcal{S}_{\tau})$ for each task-skill pair. The profile includes skill count, total skill tokens, code-token ratio, API usage, examples, validation rules, constraints, formulas, dominant skill archetype, and task domain. Based on these profiles, we define a set of information-preservation strategies $\mathcal{A}$. Each strategy specifies which class of operational anchors should remain salient after rewriting, rather than enforcing a fixed surface template. Table~\ref{tab:rewrite-strategies} summarizes the strategy families.
\begin{table}[t]
\centering
\small
\setlength{\tabcolsep}{0pt}
\renewcommand{\arraystretch}{1.05}

\begin{tabular}{
>{\raggedright\arraybackslash}p{0.24\linewidth}
@{\hspace{5pt}}
>{\raggedright\arraybackslash}p{0.32\linewidth}
@{\hspace{14pt}}
>{\raggedright\arraybackslash}p{0.34\linewidth}
}
\toprule
\textbf{Strategy} & \textbf{Preserved anchors} & \textbf{Intended use} \\
\midrule
Source-native compacting
& Original structure, local terminology, author organization
& Skills whose structure already encodes a useful workflow. \\

Workflow guarding
& Ordered steps, validation checks, constraints, pitfalls
& Tasks prone to missing checks or procedural drift. \\

API/code anchoring
& Imports, API calls, object construction, commands, snippets
& Code- or tool-heavy skills where API details prevent debugging. \\

Rule/formula anchoring
& Definitions, formulas, thresholds, schemas, conventions
& Scientific, optimization, and rule-governed tasks requiring exact invariants. \\
\bottomrule
\end{tabular}

\caption{
Skill rewriting strategies, preserved anchors, and intended use cases.
}
\label{tab:rewrite-strategies}
\end{table}



For each strategy $a$, the rewriter produces $\widetilde{s}_i^{a}=R_a(s_i)$ for every skill $s_i$. Rewrites are constrained to preserve factual content and task requirements, and are audited with lightweight checks such as token ratio, code-term coverage, code-block retention, and missing-anchor detection. When protected anchors are missing, the rewrite is repaired with a compact source-derived anchor block. This makes rewriting a controlled preservation procedure rather than unconstrained summarization.

\subsection{Task-Conditioned Strategy Learning}

Different tasks benefit from different preservation strategies, so we learn a task-conditioned strategy selector. The context is the task-skill profile $\phi_{\tau}$, the candidate actions are the final policy strategies $a\in\mathcal{A}_{\pi}\subseteq\mathcal{A}$, and the supervision signal is an economic utility $U_{\tau}^{a}$. This utility summarizes the quality--cost outcome of applying strategy $a$ to task $\tau$; its exact definition is given in Section~\ref{sec:experimental-setup}.

For each training task $\tau\in\mathcal{T}_{\mathrm{train}}$ and evaluated strategy $a$, we observe $U_{\tau}^{a}$. In our experiments, $\mathcal{T}_{\mathrm{train}}$ contains the template-calibration and policy-adaptation splits described in Section~\ref{sec:experimental-setup}; held-out tasks are excluded. We fit an action-conditioned utility model
\begin{equation}
\label{eq:value-model}
\widehat{U}_{\theta}(\phi_{\tau},a)
=
b_a+\mathbf{w}_a^{\top}z(\phi_{\tau}),
\end{equation}
where $z(\phi_{\tau})$ is the numeric feature vector derived from the task-skill profile. The parameter set $\theta=\{b_a,\mathbf{w}_a:a\in\mathcal{A}_{\pi}\}$ contains a strategy-specific intercept $b_a$ and feature weights $\mathbf{w}_a$.

The parameters are estimated by sparse utility regression:
\begingroup
\small
\begin{equation}
\label{eq:value-learning}
\begin{aligned}
\widehat{\theta}
=
\arg\min_{\theta}
&
\sum_{\tau\in\mathcal{T}_{\mathrm{train}}}
\sum_{a\in\mathcal{A}_{\tau}}
\left(
U_{\tau}^{a}
-
b_a
-
\mathbf{w}_a^{\top}z(\phi_{\tau})
\right)^2
\\
&+
\lambda_{\mathrm{sparse}}
\sum_{a\in\mathcal{A}_{\pi}}\|\mathbf{w}_a\|_1 .
\end{aligned}
\end{equation}
\endgroup
Here, $\mathcal{A}_{\tau}$ denotes the strategies evaluated for training task $\tau$. The sparsity penalty encourages the selector to rely on a small number of interpretable feature--strategy associations.

At inference time, the learned policy selects the strategy with the highest predicted utility:
\begin{equation}
\label{eq:learned-policy}
\pi_{\widehat{\theta}}(\phi_{\tau})
=
\arg\max_{a\in\mathcal{A}_{\pi}}
\widehat{U}_{\widehat{\theta}}(\phi_{\tau},a).
\end{equation}
The selected strategy is then used to rewrite the task skills. Thus, the policy chooses which class of operational anchors to preserve, rather than selecting the shortest rewrite.

\section{Experimental Setup}
\label{sec:experimental-setup}

We evaluate cost-aware skill rewriting on SkillsBench, where each task provides natural-language instructions, an executable environment, human-written skills, and an automatic verifier. Following Section~\ref{sec:method}, we rewrite only the skill documents under \texttt{environment/skills/}; task instructions, environments, and verifiers are kept fixed. This design isolates the effect of skill wording and structure from changes in the task itself, allowing differences in quality and cost to be attributed to the skill condition rather than task variation.

\subsection{Benchmark and Task Panel}
We separate unlabeled corpus profiling from performance-based evaluation. The full corpus contains 88 tasks and is used only to characterize skill structure, including skill length, number of skills, code-token ratio, validation markers, API/tool usage, formulas, and dominant archetype. This profiling stage does not use verifier scores or agent outcomes.
\begin{table}[t]
\centering
\footnotesize
\setlength{\tabcolsep}{3.0pt}
\renewcommand{\arraystretch}{1.03}
\begin{tabularx}{\columnwidth}{@{}p{0.25\columnwidth}p{0.12\columnwidth}X@{}}
\toprule
\textbf{Stage} & \textbf{Tasks} & \textbf{Use} \\
\midrule
Corpus profiling
& 88
& Static skill features only; without agent outcomes or verifier scores. \\

Runnable pool
& 86
& Tasks passing environment and verifier checks; used for execution. \\

Template calibration
& 28
& Compare fixed rewrite families and refine rewrite prompts. \\

Policy adaptation
& 38
& Learn the task-conditioned strategy selector from economic feedback. \\

Held-out evaluation
& 20
& Balanced test panel; policy and utility weights are frozen. \\

Cross-model transfer
& 86
& Frozen-policy evaluation across additional agent stacks; no feedback to training. \\
\bottomrule
\end{tabularx}
\caption{Data usage and leakage control in the evaluation protocol.}
\label{tab:data-usage}
\end{table}
For agent execution, we use 86 runnable tasks after excluding tasks with environment or verifier issues. We split this runnable pool into 28 template-calibration tasks, 38 policy-adaptation tasks, and a 20-task held-out evaluation panel. Template families and utility hyperparameters are selected only on the 28 template-calibration tasks. With these choices fixed, the final strategy-selection policy is fit using economic feedback from the template-calibration and policy-adaptation splits, then frozen before held-out and cross-model evaluation.

The held-out panel is stratified across six task families: data analysis and visualization, scientific computing, software debugging and build tasks, control and optimization, spreadsheet or office workflows, and web or visual interaction.  Appendix~\ref{app:task-splits} reports the full task list, domain labels, failure categories, and operational-section distribution of the human-written skills.

\subsection{Agent Stacks}
Our primary policy-learning and main evaluation setting is Gemini CLI with Gemini 3 Flash Preview, chosen for efficient large-scale tool-use evaluation. To test whether the learned skill-rewrite policy transfers beyond the model used for learning, we evaluate the frozen policy with three additional stacks: Gemini CLI with Gemini 3 Pro, Codex with GPT-5.4, and Claude Code with Opus 4.6. These settings cover high-throughput tool use, stronger reasoning, code-oriented file editing, and long-context agent execution.

Rewritten skill variants are generated once and cached. The same task variants are then evaluated across all agent stacks. We do not retune rewrite prompts, utility weights, or the policy for different models. Thus, cross-model comparisons measure whether the rewritten skills and frozen strategy policy transfer across agents, rather than whether the method can be specialized separately to each model. Detailed model settings, harness configurations, and aggregate run costs are reported in Appendix~\ref{app:agent-stacks}.

\subsection{Skill Conditions}

We compare original human-written skills (SkillsBench baseline), fixed rewrite strategies, and policy-selected rewrites. The fixed strategies are API/code anchoring, rule/formula anchoring, and workflow guarding, corresponding to the preservation families in Table~\ref{tab:rewrite-strategies}. Source-native compacting is included only in diagnostic analyses because pilot runs showed weaker validity and quality. Thus, the policy action space is $\mathcal{A}_{\pi}=\{\textsc{API/code},\textsc{Rule/formula},\textsc{Workflow}\}$. The policy-selected condition uses the task-conditioned strategy learner from Section~\ref{sec:method}. Utility weights are fixed after template calibration; the selector is then trained on calibration and adaptation tasks and frozen for held-out and cross-model evaluation.
\subsection{Metrics and Utility}

We report both task quality and execution economy. A run is considered valid if the environment launches, the agent completes the attempt, the verifier executes, and token logs are recorded. Infrastructure, harness, or verifier-execution failures are excluded from metric averages and counted separately; agent failures on the task itself remain valid runs and receive the verifier score returned by the task.

When multiple trials are available for the same task, agent stack, and skill condition, we first average over valid trials within that task. We then macro-average the resulting task-level values across tasks, so that each task has equal weight. The Valid column reports the number of tasks with at least one valid trial under the corresponding condition.

For cost, we separately count direct skill tokens and downstream agent tokens. Direct skill tokens are computed from rewritten \texttt{SKILL.md} files, while downstream cost is computed from API input and output token logs collected during agent execution. For each agent stack and task, cost ratios are normalized by the original-skill baseline, denoted by $a=0$, before macro-averaging across tasks.

Let $q_{\tau}^{a}$ be the verifier score, $C_{\mathrm{skill},\tau}^{a}$ the direct skill-token cost, $C_{\mathrm{agent},\tau}^{a}$ the downstream agent-token cost, and $C_{\tau}^{a}=C_{\mathrm{skill},\tau}^{a}+C_{\mathrm{agent},\tau}^{a}$ the total cost. We report quality retention and total cost ratio:
\begin{equation}
\label{eq:quality-cost-ratio}
\mathrm{QR}_{\tau}^{a}
=
\frac{q_{\tau}^{a}}{\max(q_{\tau}^{0},\epsilon)},
\quad
\rho_{\tau}^{a}
=
\frac{C_{\tau}^{a}}{\max(C_{\tau}^{0},\epsilon)}.
\end{equation}
We also report direct skill-token and downstream agent-token ratios:
\begin{equation}
\label{eq:component-ratios}
\begin{aligned}
r_{s,\tau}^{a}
&=
\frac{C_{\mathrm{skill},\tau}^{a}}
{\max(C_{\mathrm{skill},\tau}^{0},\epsilon)}, \\
r_{a,\tau}^{a}
&=
\frac{C_{\mathrm{agent},\tau}^{a}}
{\max(C_{\mathrm{agent},\tau}^{0},\epsilon)}.
\end{aligned}
\end{equation}
The execution-cost change is $\Delta_{\tau}^{a}=r_{a,\tau}^{a}-1$, and the execution-overrun penalty is $\mathrm{EO}_{\tau}^{a}=\max(0,\Delta_{\tau}^{a})$. Thus, $\rho<1$ indicates lower total cost, while $\Delta>0$ indicates that rewritten skills cause more downstream agent-token usage than the original skills.

We define near-lossless dividend as
\begin{equation}
\label{eq:near-lossless-dividend}
\mathrm{NLD}_{\tau}^{a}
=
\mathbb{I}
\bigl[\mathrm{QR}_{\tau}^{a}\ge 1-\delta\bigr]
\max(0,1-\rho_{\tau}^{a}),
\end{equation}
which rewards cost savings when quality remains within tolerance $\delta$ of the original-skill baseline.

For strategy learning, we use the scalar economic utility
\begin{equation}
\label{eq:utility}
\begin{aligned}
U_{\tau}^{a}
&=
\mathrm{QR}_{\tau}^{a}
+
\lambda_{\mathrm{save}}(1-\rho_{\tau}^{a})
\\
&\quad
-
\lambda_{\mathrm{over}}\mathrm{EO}_{\tau}^{a}
+
\lambda_{\mathrm{nld}}\mathrm{NLD}_{\tau}^{a}.
\end{aligned}
\end{equation}

The utility rewards quality retention, total cost reduction, and near-lossless savings, while penalizing downstream execution cost that exceeds the original-skill baseline. The utility is an experimental selection objective for this benchmark rather than a universal measure of skill quality. We choose the utility weights and tolerance $\delta$ using only the 28 template-calibration tasks; they are then fixed before policy adaptation, held-out evaluation, and cross-model transfer. Nonlinear quantities such as QR, $\rho$, EO, NLD, and $U$ are computed from task-level aggregates and then macro-averaged across tasks. Exact values and selection details are given in Appendix~\ref{app:policy-details}. Full token logs and auxiliary economic diagnostics are reported in Appendix~\ref{app:economic-metrics}.

\section{Results and Analysis}

\label{sec:results}

\subsection{Main Quality--Cost Results}
\begin{table*}[t]
\centering
\footnotesize
\setlength{\tabcolsep}{3.0pt}
\renewcommand{\arraystretch}{1.03}

\begin{tabular}{@{}llrrrrrrrrrr@{}}
\toprule
\textbf{Setting} &
\textbf{Skill condition} &
\textbf{Valid} &
\textbf{Partial} &
\textbf{Reward} &
\textbf{QR} &
\boldmath$r_s$ &
\boldmath$r_a$ &
\boldmath$\rho$ &
\boldmath$\mathrm{EO}$ &
\boldmath$\Delta$ &
\textbf{NLD} \\
\midrule

\multicolumn{12}{@{}l}{\textit{Held-out evaluation: 20-task balanced panel}} \\
Gemini-CLI + Gemini 3 Flash
& Original skills
& 18/20 & \second{0.815} & \second{0.61} & \second{1.00} & 1.00 & 1.00 & 1.00 & 0.00 & 0.00 & 0.00 \\
& API/code anchored
& 17/20 & 0.792 & 0.59 & 0.97 & \second{0.60} & \second{0.96} & 0.94 & 0.00 & -0.04 & \second{0.04} \\
& Rule/formula anchored
& 19/20 & 0.555 & 0.42 & 0.68 & 0.65 & \best{0.94} & \best{0.92} & 0.00 & \best{-0.06} & 0.00 \\
& Workflow-guarded
& 19/20 & 0.731 & 0.55 & 0.90 & \best{0.58} & 1.14 & 1.11 & 0.14 & +0.14 & 0.00 \\
\oursrow & Policy-selected
& 19/20 & \best{0.819} & \best{0.63} & \best{1.01} & 0.62 & \best{0.94} & \second{0.93} & 0.00 & \best{-0.06} & \best{0.07} \\
\midrule

\multicolumn{12}{@{}l}{\textit{Frozen-policy transfer: 86 runnable tasks}} \\
Gemini-CLI + Gemini 3 Pro
& Original skills
& 72/86 & 0.792 & 0.56 & 1.00 & 1.00 & 1.00 & 1.00 & 0.00 & 0.00 & 0.00 \\
& API/code anchored
& 73/86 & \second{0.808} & \second{0.58} & \second{1.02} & \second{0.61} & 0.92 & 0.91 & 0.00 & -0.08 & \second{0.08} \\
& Rule/formula anchored
& 74/86 & 0.704 & 0.46 & 0.89 & 0.66 & \second{0.89} & \second{0.88} & 0.00 & -0.11 & 0.00 \\
& Workflow-guarded
& 73/86 & 0.778 & 0.54 & 0.98 & \best{0.59} & 1.07 & 1.04 & 0.07 & +0.07 & 0.00 \\
\oursrow & Policy-selected
& 75/86 & \best{0.822} & \best{0.61} & \best{1.04} & 0.63 & \best{0.88} & \best{0.87} & 0.00 & \best{-0.12} & \best{0.12} \\
\midrule

Codex + GPT-5.4
& Original skills
& 74/86 & 0.835 & 0.62 & 1.00 & 1.00 & 1.00 & 1.00 & 0.00 & 0.00 & 0.00 \\
& API/code anchored
& 76/86 & \second{0.852} & \second{0.65} & \second{1.02} & \second{0.61} & 0.89 & 0.88 & 0.00 & -0.11 & \second{0.11} \\
& Rule/formula anchored
& 77/86 & 0.732 & 0.50 & 0.88 & 0.66 & \second{0.86} & \second{0.85} & 0.00 & -0.14 & 0.00 \\
& Workflow-guarded
& 75/86 & 0.810 & 0.59 & 0.97 & \best{0.59} & 1.03 & 1.00 & 0.03 & +0.03 & 0.00 \\
\oursrow & Policy-selected
& 78/86 & \best{0.862} & \best{0.68} & \best{1.03} & 0.63 & \best{0.84} & \best{0.83} & 0.00 & \best{-0.16} & \best{0.16} \\
\midrule

Claude Code + Opus 4.6
& Original skills
& 76/86 & 0.852 & 0.65 & 1.00 & 1.00 & 1.00 & 1.00 & 0.00 & 0.00 & 0.00 \\
& API/code anchored
& 77/86 & \second{0.864} & \second{0.67} & \second{1.01} & \second{0.61} & 0.91 & 0.90 & 0.00 & -0.09 & \second{0.10} \\
& Rule/formula anchored
& 78/86 & 0.756 & 0.53 & 0.89 & 0.66 & \second{0.88} & \second{0.87} & 0.00 & -0.12 & 0.00 \\
& Workflow-guarded
& 78/86 & 0.832 & 0.63 & 0.98 & \best{0.59} & 1.05 & 1.02 & 0.05 & +0.05 & 0.00 \\
\oursrow & Policy-selected
& 79/86 & \best{0.874} & \best{0.70} & \best{1.03} & 0.63 & \best{0.87} & \best{0.86} & 0.00 & \best{-0.13} & \best{0.14} \\
\bottomrule
\end{tabular}

\caption{
Main quality--cost results.
The first block reports held-out evaluation on the 20-task balanced panel.
The remaining blocks report frozen-policy transfer on the 86 runnable tasks.
Partial and Reward are task-level macro-averages over valid trials; Valid counts tasks with at least one valid trial.
Cost ratios are normalized to the original-skill baseline within each task and agent stack before macro-averaging.
$\Delta=r_a-1$ measures downstream execution-cost change.
Bold and underlined values mark the best and second-best values within each block.
}
\label{tab:main-results}
\end{table*}

Table~\ref{tab:main-results} shows that policy-selected rewriting improves the quality--cost trade-off. In the primary 20-task held-out setting, the frozen policy preserves verifier quality while reducing direct skill-token cost ($r_s=0.62$), downstream agent-token cost ($r_a=0.94$), and total cost ($\rho=0.93$), corresponding to 7.0\% total-cost and 6.0\% downstream-token savings. Under frozen transfer to Gemini Pro, Codex, and Claude stacks, the corresponding reductions average 14.7\% and 13.7\%; we treat these full-pool results as robustness evidence because some tasks were used during calibration or policy adaptation.

The fixed strategies explain why skill rewriting cannot be reduced to choosing the shortest text. API/code anchoring gives a strong quality--cost balance, rule/formula anchoring is economical but loses quality when applied broadly, and workflow guarding shortens skills while increasing downstream execution cost. The selector improves the frontier by choosing among preservation objectives instead of committing to one template.

The fixed strategies expose why skill rewriting cannot be reduced to choosing the shortest text. API/code anchoring gives a strong quality--cost balance on the held-out panel, rule/formula anchoring is economical but loses substantial quality when applied broadly, and workflow guarding shortens skills while increasing downstream execution cost. The selector improves the frontier by choosing among preservation objectives instead of committing to one template. Cross-model results show the same qualitative pattern under different agent stacks: policy-selected rewrites consistently reduce normalized total cost while maintaining or improving mean verifier scores.

\subsection{Execution-Cost Trade-offs of Fixed Templates}
\begin{figure}[t]
    \centering
    \includegraphics[width=\linewidth]{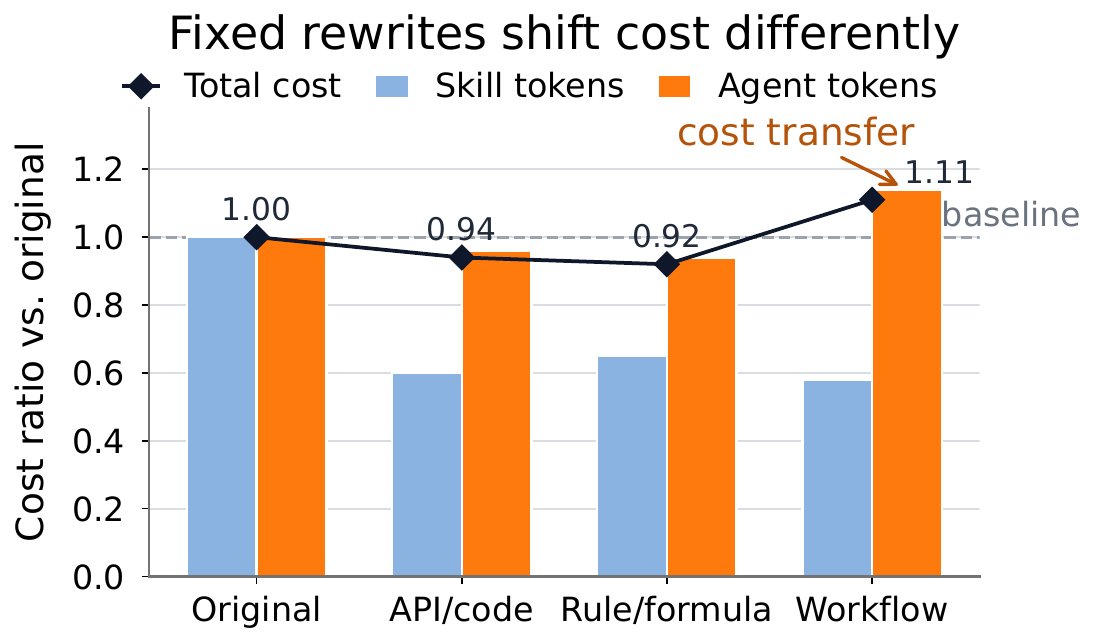}
    \caption{
Cost decomposition for fixed rewrite strategies in the held-out setting.
Bars show direct skill-token ratio $r_s$ and downstream agent-token ratio $r_a$; diamonds show total cost ratio $\rho$.
Workflow guarding reduces skill length but increases downstream execution cost.
    }
    \label{fig:fixed-template-cost-transfer}
\end{figure}

Figure~\ref{fig:fixed-template-cost-transfer} decomposes fixed-strategy costs in the primary held-out setting. API/code anchoring reduces both direct skill tokens and downstream agent tokens, whereas workflow guarding reduces skill length but raises downstream usage to $1.14\times$ the original-skill baseline. This illustrates why skill rewriting cannot be evaluated by prompt length alone: shorter skills can still make the agent spend more tokens during execution.

\subsection{What Does the Policy Learn?}
\begin{figure}[t]
  \centering

  \includegraphics[width=1\linewidth]{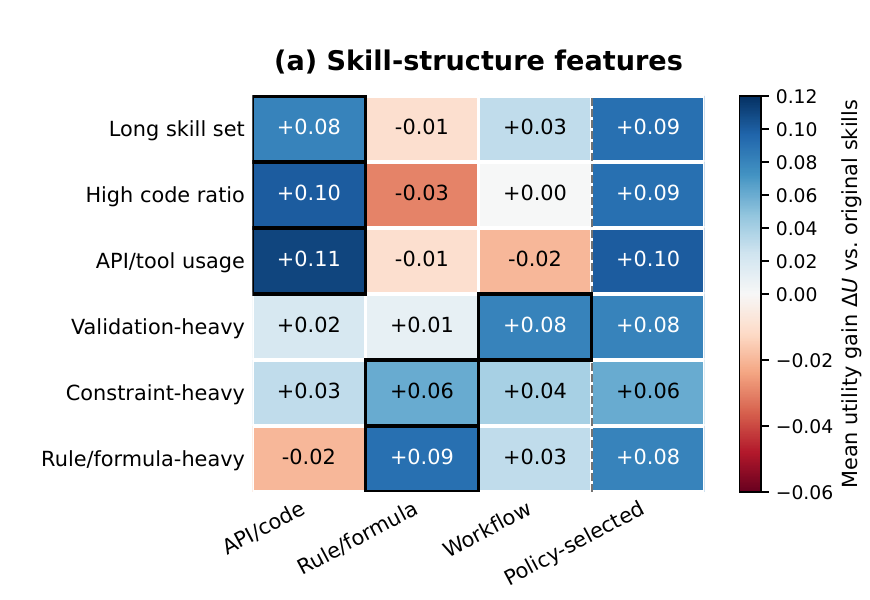}

  \vspace{0.3em}

  \includegraphics[width=1\linewidth]{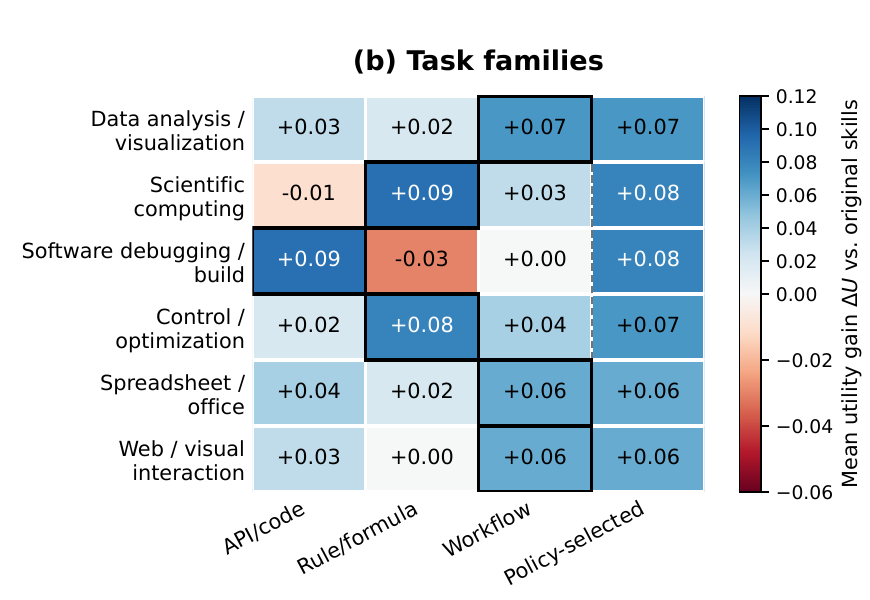}

  \caption{
Strategy preferences by skill structure and task family.
Cells show mean utility gain $\Delta U=U^a-U^0$.
Black boxes mark the best fixed strategy in each row.
The policy-selected column evaluates the frozen selector, not hindsight selection, so it may be slightly below the best fixed strategy for a group.
}
  \label{fig:strategy-preferences}
\end{figure}
We analyze which preservation strategies are favored by different task and skill structures. For each held-out group, Figure~\ref{fig:strategy-preferences} reports the mean utility gain $\Delta U=U^{a}-U^{0}$ relative to original skills. The left panel groups tasks by skill-structure features, while the right panel groups them by task family; these groups are not mutually exclusive.

The heatmaps show that strategy preferences are structured rather than random. API/code anchoring is most useful for long, code-heavy, and API/tool-using skills; rule/formula anchoring is strongest when exact definitions, schemas, units, or scientific conventions dominate; and workflow guarding is useful only when procedural checks outweigh its tendency to lengthen downstream trajectories. The policy-selected column tests whether these preferences can be captured by a frozen selector rather than by choosing the best strategy in hindsight. Its positive utility across groups supports the main claim: economically effective skill rewriting depends on preserving task-relevant procedural anchors, not merely compressing skill text.

\subsection{Ablations and Robustness}

We ablate the components that connect rewriting to economic strategy learning. Table~\ref{tab:ablation-robustness} examines which components of the policy objective are responsible for the quality--cost trade-off. The ablation block uses the primary held-out Gemini Flash setting. A quality-only selector obtains slightly higher verifier score, but loses the cost advantage because it allows downstream execution to increase. Removing the overrun penalty has a similar effect: direct skill cost is reduced, but the agent spends more tokens after the prompt. Removing the near-lossless dividend makes the selector more aggressively cost-seeking and slightly reduces quality retention. Removing anchor repair causes the largest quality drop, confirming that protected implementation and rule anchors are necessary for reliable rewriting.

\begin{table}[t]
\centering
\footnotesize
\setlength{\tabcolsep}{3.1pt}
\renewcommand{\arraystretch}{1.02}
\begin{tabular}{@{}lrrrrr@{}}
\toprule
\textbf{Variant / setting} &
\textbf{Part.} &
\textbf{QR} &
\boldmath$\rho$ &
\boldmath$\Delta$ &
\textbf{NLD} \\
\midrule
\multicolumn{6}{@{}l}{\textit{Ablations on primary stack}} \\
\oursrow Full policy
& \second{0.819} & \second{1.01} & 0.93 & \second{-0.06} & \best{0.07} \\
Quality-only
& \best{0.826} & \best{1.02} & 1.03 & +0.08 & 0.01 \\
No overrun penalty
& 0.817 & 1.00 & 0.99 & +0.03 & 0.02 \\
No NLD bonus
& 0.802 & 0.98 & \best{0.91} & \best{-0.07} & 0.03 \\
No anchor repair
& 0.776 & 0.95 & \second{0.92} & -0.05 & 0.02 \\
Fixed API/code
& 0.792 & 0.97 & 0.94 & -0.04 & \second{0.04} \\
\midrule
\multicolumn{6}{@{}l}{\textit{Frozen-policy transfer}} \\
Gemini Flash
& 0.819 & 1.01 & 0.93 & -0.06 & 0.07 \\
Gemini Pro
& 0.822 & \best{1.04} & 0.87 & -0.12 & 0.12 \\
Codex + GPT-5.4
& \second{0.862} & \second{1.03} & \best{0.83} & \best{-0.16} & \best{0.16} \\
Claude + Opus 4.6
& \best{0.874} & \second{1.03} & \second{0.86} & \second{-0.13} & \second{0.14} \\
\bottomrule
\end{tabular}
\caption{
Ablations and frozen-policy robustness.
The ablation block is evaluated on the primary held-out Gemini Flash setting.
The transfer block reports the policy-selected rows from Table~\ref{tab:main-results}.
Part. is verifier partial score; QR is quality retention; $\rho$ is total cost ratio; $\Delta=r_a-1$ is downstream execution-cost change; and NLD is near-lossless dividend.
}
\label{tab:ablation-robustness}
\end{table}

The transfer block reports the policy-selected rows from Table~\ref{tab:main-results}. The same frozen policy reduces total cost across all four reported settings, with $\rho$ ranging from 0.83 to 0.93 and no positive downstream overrun. This suggests that the selector captures reusable structure in skill rewriting rather than a behavior tied to one agent stack.

\section{Conclusion}
We studied skill rewriting as cost-aware operational knowledge preservation. Instead of treating skills as prompts to be shortened, we asked which procedural anchors should be retained because they shape both task quality and downstream execution. Our framework profiles task-skill structure, rewrites skills with preservation-oriented strategies, evaluates them under fixed task conditions, and distills a task-conditioned selector. Experiments on SkillsBench show that fixed strategies expose complementary quality--cost trade-offs, and that a frozen selector can reduce execution cost in held-out and cross-stack evaluations while preserving verifier quality. These results suggest that effective skill design is not prompt compression, but cost-aware knowledge engineering for agent execution.

\section*{Limitations}

This work studies cost-aware rewriting of textual agent skill documents under controlled benchmark conditions. We keep task instructions, execution environments, and verifiers fixed, which allows us to isolate the effect of skill rewriting, but also means that the experiments do not cover all forms of agent knowledge, such as dynamically retrieved resources or continuously updated deployment-time skills.

Our cost analysis focuses on token-based measurements, including direct skill-token cost and downstream agent-token cost. This provides a clear and reproducible view of execution economy, but real deployment cost may also depend on latency, provider-specific pricing, caching, hardware, and human review. Future work can extend the same evaluation framework to these broader cost factors and to additional agent environments.

\section*{Ethical Considerations}

This work studies the rewriting of agent skill documents for cost-aware execution. The experiments are conducted on benchmark tasks with executable environments and automatic verifiers, and do not involve human subjects, private user data, or personally identifiable information. We rewrite only the skill documents while keeping task instructions, environments, and verifiers fixed, which limits the scope of the intervention to the procedural knowledge supplied to the agent.

A potential risk of skill rewriting is that compression or restructuring may remove safety-relevant constraints, validation checks, failure-handling instructions, or ``do not'' rules that are important for reliable agent behavior. This risk is directly related to the main argument of the paper: effective rewriting should preserve task-relevant operational anchors rather than merely shorten text. Our rewriting protocol therefore includes preservation-oriented strategies and post-generation audits for code/API terms, validation rules, code-block retention, and missing-anchor repair. Nevertheless, these checks are lightweight and should not be treated as a substitute for human review in safety-critical, legal, medical, financial, or high-stakes deployment settings.

Another consideration is that reducing agent execution cost can make agent systems easier to deploy at scale. While lower token usage may reduce computational expense and environmental footprint, it may also encourage broader use of automated agents in settings where reliability, accountability, or human oversight is insufficient. Our results should therefore be interpreted as evidence about quality--cost trade-offs in controlled benchmark settings, not as a blanket recommendation to replace careful skill design, monitoring, or domain-specific validation.

Finally, the work uses existing benchmark artifacts and commercial or research agent stacks for evaluation. Any release of rewritten skill variants, logs, or tooling should respect the licenses and terms of the underlying benchmark, model providers, and agent frameworks. We encourage reporting token usage, runtime, and failure cases alongside quality metrics so that future comparisons of agent skills reflect both performance and resource costs.

\bibliography{custom}

@inproceedings{yao2023react,
  title = {{ReAct}: Synergizing Reasoning and Acting in Language Models},
  author = {Yao, Shunyu and Zhao, Jeffrey and Yu, Dian and Du, Nan and Shafran, Izhak and Narasimhan, Karthik and Cao, Yuan},
  booktitle = {International Conference on Learning Representations},
  year = {2023},
  url = {https://openreview.net/forum?id=WE_vluYUL-X}
}

@inproceedings{liu2024agentbench,
  title = {{AgentBench}: Evaluating {LLM}s as Agents},
  author = {Liu, Xiao and Yu, Hao and Zhang, Hanchen and Xu, Yifan and Lei, Xuanyu and Lai, Hanyu and Gu, Yu and Ding, Hangliang and Men, Kaiwen and Yang, Kejuan and Zhang, Shudan and Deng, Xiang and Zeng, Aohan and Du, Zhengxiao and Zhang, Chenhui and Shen, Sheng and Zhang, Tianjun and Su, Yu and Sun, Huan and Huang, Minlie and Dong, Yuxiao and Tang, Jie},
  booktitle = {International Conference on Learning Representations},
  year = {2024},
  url = {https://openreview.net/forum?id=zAdUB0aCTQ}
}

@inproceedings{zhou2024webarena,
  title = {{WebArena}: A Realistic Web Environment for Building Autonomous Agents},
  author = {Zhou, Shuyan and Xu, Frank F. and Zhu, Hao and Zhou, Xuhui and Lo, Robert and Sridhar, Abishek and Cheng, Xianyi and Ou, Tianyue and Bisk, Yonatan and Fried, Daniel and Alon, Uri and Neubig, Graham},
  booktitle = {International Conference on Learning Representations},
  year = {2024},
  url = {https://openreview.net/forum?id=oKn9c6ytLx}
}

@inproceedings{jimenez2024swebench,
  title = {{SWE}-bench: Can Language Models Resolve Real-World GitHub Issues?},
  author = {Jimenez, Carlos E. and Yang, John and Wettig, Alexander and Yao, Shunyu and Pei, Kexin and Press, Ofir and Narasimhan, Karthik},
  booktitle = {International Conference on Learning Representations},
  year = {2024},
  url = {https://openreview.net/forum?id=VTF8yNQM66}
}

@inproceedings{yao2024taubench,
  title = {$\tau$-bench: A Benchmark for Tool-Agent-User Interaction in Real-World Domains},
  author = {Yao, Shunyu and Shinn, Noah and Razavi, Pedram and Narasimhan, Karthik},
  booktitle = {International Conference on Learning Representations},
  year = {2025},
  url = {https://openreview.net/forum?id=roNSXZpUDN}
}

@inproceedings{yang2024sweagent,
  title = {{SWE}-agent: Agent-Computer Interfaces Enable Automated Software Engineering},
  author = {Yang, John and Jimenez, Carlos E. and Wettig, Alexander and Lieret, Kilian and Yao, Shunyu and Narasimhan, Karthik and Press, Ofir},
  booktitle = {Advances in Neural Information Processing Systems},
  year = {2024},
  url = {https://openreview.net/forum?id=mXpq6ut8J3}
}

@article{wang2023voyager,
  title = {Voyager: An Open-Ended Embodied Agent with Large Language Models},
  author = {Wang, Guanzhi and Xie, Yuqi and Jiang, Yunfan and Mandlekar, Ajay and Xiao, Chaowei and Zhu, Yuke and Fan, Linxi and Anandkumar, Anima},
  journal = {Transactions on Machine Learning Research},
  year = {2024},
  url = {https://openreview.net/forum?id=ehfRiF0R3a}
}

@inproceedings{jiang2023llmlingua,
  title = {{LLMLingua}: Compressing Prompts for Accelerated Inference of Large Language Models},
  author = {Jiang, Huiqiang and Wu, Qianhui and Lin, Chin-Yew and Yang, Yuqing and Qiu, Lili},
  booktitle = {Proceedings of the 2023 Conference on Empirical Methods in Natural Language Processing},
  pages = {13358--13376},
  year = {2023},
  address = {Singapore},
  publisher = {Association for Computational Linguistics},
  doi = {10.18653/v1/2023.emnlp-main.825},
  url = {https://aclanthology.org/2023.emnlp-main.825/}
}

@inproceedings{jiang2024longllmlingua,
  title = {{LongLLMLingua}: Accelerating and Enhancing {LLM}s in Long Context Scenarios via Prompt Compression},
  author = {Jiang, Huiqiang and Wu, Qianhui and Luo, Xufang and Li, Dongsheng and Lin, Chin-Yew and Yang, Yuqing and Qiu, Lili},
  booktitle = {Proceedings of the 62nd Annual Meeting of the Association for Computational Linguistics},
  pages = {1658--1677},
  year = {2024},
  address = {Bangkok, Thailand},
  publisher = {Association for Computational Linguistics},
  doi = {10.18653/v1/2024.acl-long.91},
  url = {https://aclanthology.org/2024.acl-long.91/}
}

@inproceedings{pryzant2023automatic,
  title = {Automatic Prompt Optimization with ``Gradient Descent'' and Beam Search},
  author = {Pryzant, Reid and Iter, Dan and Li, Jerry and Lee, Yin and Zhu, Chenguang and Zeng, Michael},
  booktitle = {Proceedings of the 2023 Conference on Empirical Methods in Natural Language Processing},
  pages = {7957--7968},
  year = {2023},
  address = {Singapore},
  publisher = {Association for Computational Linguistics},
  doi = {10.18653/v1/2023.emnlp-main.494},
  url = {https://aclanthology.org/2023.emnlp-main.494/}
}

@misc{fernando2023promptbreeder,
  title = {{Promptbreeder}: Self-Referential Self-Improvement via Prompt Evolution},
  author = {Fernando, Chrisantha and Banarse, Dylan Sunil and Michalewski, Henryk and Osindero, Simon and Rockt{"a}schel, Tim},
  year = {2023},
  url = {https://openreview.net/forum?id=HKkiX32Zw1},
  note = {Submitted to ICLR 2024}
}

@inproceedings{khattab2024dspy,
  title = {{DSPy}: Compiling Declarative Language Model Calls into State-of-the-Art Pipelines},
  author = {Khattab, Omar and Singhvi, Arnav and Maheshwari, Paridhi and Zhang, Zhiyuan and Santhanam, Keshav and {Sri Vardhamanan A} and Haq, Saiful and Sharma, Ashutosh and Joshi, Thomas T. and Moazam, Hanna and Miller, Heather and Zaharia, Matei and Potts, Christopher},
  booktitle = {International Conference on Learning Representations},
  year = {2024},
  url = {https://proceedings.iclr.cc/paper_files/paper/2024/hash/f1cf02ce09757f57c3b93c0db83181e0-Abstract-Conference.html}
}

@misc{anthropic2025agentskills,
  author = {{Anthropic}},
  title = {Equipping Agents for the Real World with Agent Skills},
  year = {2025},
  howpublished = {\url{https://www.anthropic.com/engineering/equipping-agents-for-the-real-world-with-agent-skills}},
  note = {Accessed: 2026-05-17}
}

@misc{openai2026skills,
  author = {{OpenAI}},
  title = {Skills in ChatGPT},
  year = {2026},
  howpublished = {\url{https://help.openai.com/en/articles/20001066-skills-in-chatgpt}},
  note = {Accessed: 2026-05-17}
}

@inproceedings{schick2023toolformer,
  title = {{Toolformer}: Language Models Can Teach Themselves to Use Tools},
  author = {Schick, Timo and Dwivedi-Yu, Jane and Dessi, Roberto and Raileanu, Roberta and Lomeli, Maria and Hambro, Eric and Zettlemoyer, Luke and Cancedda, Nicola and Scialom, Thomas},
  booktitle = {Advances in Neural Information Processing Systems},
  volume = {36},
  year = {2023},
  url = {https://openreview.net/forum?id=Yacmpz84TH}
}

@inproceedings{qin2024toolllm,
  title = {{ToolLLM}: Facilitating Large Language Models to Master 16000+ Real-World {API}s},
  author = {Qin, Yujia and Liang, Shihao and Ye, Yining and Zhu, Kunlun and Yan, Lan and Lu, Yaxi and Lin, Yankai and Cong, Xin and Tang, Xiangru and Qian, Bill and Zhao, Sihan and Hong, Lauren and Tian, Runchu and Xie, Ruobing and Zhou, Jie and Gerstein, Mark and Li, Dahai and Liu, Zhiyuan and Sun, Maosong},
  booktitle = {International Conference on Learning Representations},
  year = {2024},
  url = {https://openreview.net/forum?id=dHng2O0Jjr}
}

@inproceedings{li2023apibank,
  title = {{API}-Bank: A Comprehensive Benchmark for Tool-Augmented {LLM}s},
  author = {Li, Minghao and Zhao, Yingxiu and Yu, Bowen and Song, Feifan and Li, Hangyu and Yu, Haiyang and Li, Zhoujun and Huang, Fei and Li, Yongbin},
  booktitle = {Proceedings of the 2023 Conference on Empirical Methods in Natural Language Processing},
  pages = {3102--3116},
  year = {2023},
  address = {Singapore},
  publisher = {Association for Computational Linguistics},
  doi = {10.18653/v1/2023.emnlp-main.187},
  url = {https://aclanthology.org/2023.emnlp-main.187/}
}

@article{xu2026agentskills,
  title = {Agent Skills for Large Language Models: Architecture, Acquisition, Security, and the Path Forward},
  author = {Xu, Renjun and Yan, Yang},
  journal = {arXiv preprint arXiv:2602.12430},
  year = {2026},
  note = {Accepted by Agent Skills '26 Workshop at ACM Conference on AI and Agentic Systems 2026},
  url = {https://arxiv.org/abs/2602.12430}
}

@article{chen2023frugalgpt,
  title = {{FrugalGPT}: How to Use Large Language Models While Reducing Cost and Improving Performance},
  author = {Chen, Lingjiao and Zaharia, Matei and Zou, James},
  journal = {arXiv preprint arXiv:2305.05176},
  year = {2023},
  url = {https://arxiv.org/abs/2305.05176}
}

@inproceedings{yue2024large,
  title = {Large Language Model Cascades with Mixture of Thought Representations for Cost-Efficient Reasoning},
  author = {Yue, Murong and Zhao, Jie and Zhang, Min and Du, Liang and Yao, Ziyu},
  booktitle = {International Conference on Learning Representations},
  year = {2024},
  url = {https://openreview.net/forum?id=6okaSfANzh}
}

@inproceedings{ge2024icae,
  title = {In-context Autoencoder for Context Compression in a Large Language Model},
  author = {Ge, Tao and Hu, Jing and Wang, Lei and Wang, Xun and Chen, Si-Qing and Wei, Furu},
  booktitle = {International Conference on Learning Representations},
  year = {2024},
  url = {https://openreview.net/forum?id=uREj4ZuGJE}
}

@misc{agentskills2026standard,
  title = {Agent Skills Overview},
  author = {{Agent Skills}},
  year = {2026},
  howpublished = {\url{https://agentskills.io/home}},
  note = {Accessed: 2026-05-17}
}

@inproceedings{pan2024llmlingua2,
  title = {{LLMLingua}-2: Data Distillation for Efficient and Faithful Task-Agnostic Prompt Compression},
  author = {Pan, Zhuoshi and Wu, Qianhui and Jiang, Huiqiang and Xia, Menglin and Luo, Xufang and Zhang, Jue and Lin, Qingwei and R{\"u}hle, Victor and Yang, Yuqing and Lin, Chin-Yew and Zhao, H. Vicky and Qiu, Lili and Zhang, Dongmei},
  booktitle = {Findings of the Association for Computational Linguistics: ACL 2024},
  pages = {963--981},
  year = {2024},
  address = {Bangkok, Thailand},
  publisher = {Association for Computational Linguistics},
  doi = {10.18653/v1/2024.findings-acl.57},
  url = {https://aclanthology.org/2024.findings-acl.57/}
}

@inproceedings{li2025promptcompressionsurvey,
  title = {Prompt Compression for Large Language Models: A Survey},
  author = {Li, Zongqian and Liu, Yinhong and Su, Yixuan and Collier, Nigel},
  booktitle = {Proceedings of the 2025 Conference of the Nations of the Americas Chapter of the Association for Computational Linguistics: Human Language Technologies},
  pages = {7182--7195},
  year = {2025},
  address = {Albuquerque, New Mexico},
  publisher = {Association for Computational Linguistics},
  doi = {10.18653/v1/2025.naacl-long.368},
  url = {https://aclanthology.org/2025.naacl-long.368/}
}

@article{shirkavand2025costaware,
  title = {Cost-Aware Contrastive Routing for {LLM}s},
  author = {Shirkavand, Reza and Gao, Shangqian and Yu, Peiran and Huang, Heng},
  journal = {arXiv preprint arXiv:2508.12491},
  year = {2025},
  url = {https://arxiv.org/abs/2508.12491}
}

@article{xiao2026agentdiet,
  title = {Reducing Cost of {LLM} Agents with Trajectory Reduction},
  author = {Xiao, Yuan-An and Gao, Pengfei and Peng, Chao and Xiong, Yingfei},
  journal = {Proceedings of the ACM on Software Engineering},
  volume = {3},
  number = {FSE},
  articleno = {FSE056},
  year = {2026},
  doi = {10.1145/3797084},
  url = {https://doi.org/10.1145/3797084}
}

@misc{jiang2026agenticskills,
  title = {{SoK}: Agentic Skills -- Beyond Tool Use in {LLM} Agents},
  author = {Jiang, Yanna and Li, Delong and Deng, Haiyu and Ma, Baihe and Wang, Xu and Wang, Qin and Yu, Guangsheng},
  year = {2026},
  eprint = {2602.20867},
  archivePrefix = {arXiv},
  primaryClass = {cs.AI},
  url = {https://arxiv.org/abs/2602.20867}
}

@misc{li2026skillsbench,
  title = {{SkillsBench}: Benchmarking How Well Agent Skills Work Across Diverse Tasks},
  author = {Li, Xiangyi and Chen, Wenbo and Liu, Yimin and Zheng, Shenghan and Chen, Xiaokun and He, Yifeng and Li, Yubo and You, Bingran and Shen, Haotian and Sun, Jiankai and Wang, Shuyi and Zeng, Qunhong and Wang, Di and Zhao, Xuandong and Wang, Yuanli and Ben Chaim, Roey and Di, Zonglin and Gao, Yipeng and He, Junwei and He, Yizhuo and Jing, Liqiang and Kong, Luyang and Lan, Xin and Li, Jiachen and Li, Songlin and Li, Yijiang and Lin, Yueqian and Liu, Xinyi and Liu, Xuanqing and Lyu, Haoran and Ma, Ze and Wang, Bowei and Wang, Runhui and Wang, Tianyu and Ye, Wengao and Zhang, Yue and Xing, Hanwen and Xue, Yiqi and Dillmann, Steven and Lee, Han-chung},
  year = {2026},
  eprint = {2602.12670},
  archivePrefix = {arXiv},
  primaryClass = {cs.AI},
  url = {https://arxiv.org/abs/2602.12670}
}

@misc{zhou2026agentskillssurvey,
  title = {A Comprehensive Survey on Agent Skills: Taxonomy, Techniques, and Applications},
  author = {Zhou, Yingli and Shu, Wang and Su, Yaodong and Du, Wenchuan and Fang, Yixiang and Lin, Xuemin},
  year = {2026},
  eprint = {2605.07358},
  archivePrefix = {arXiv},
  primaryClass = {cs.IR},
  url = {https://arxiv.org/abs/2605.07358}
}

@misc{liu2026skillswild,
  title = {How Well Do Agentic Skills Work in the Wild: Benchmarking {LLM} Skill Usage in Realistic Settings},
  author = {Liu, Yujian and Ji, Jiabao and An, Li and Jaakkola, Tommi and Zhang, Yang and Chang, Shiyu},
  year = {2026},
  eprint = {2604.04323},
  archivePrefix = {arXiv},
  primaryClass = {cs.CL},
  url = {https://arxiv.org/abs/2604.04323}
}

@misc{zhong2026skilllearnbench,
  title = {{SkillLearnBench}: Benchmarking Continual Learning Methods for Agent Skill Generation on Real-World Tasks},
  author = {Zhong, Shanshan and Lu, Yi and Ning, Jingjie and Wan, Yibing and Feng, Lihan and Ao, Yuyi and Ribeiro, Leonardo F. R. and Dreyer, Markus and Ammirati, Sean and Xiong, Chenyan},
  year = {2026},
  eprint = {2604.20087},
  archivePrefix = {arXiv},
  primaryClass = {cs.AI},
  url = {https://arxiv.org/abs/2604.20087}
}

@inproceedings{ong2024routellm,
  title = {{RouteLLM}: Learning to Route {LLM}s from Preference Data},
  author = {Ong, Isaac and Almahairi, Amjad and Wu, Vincent and Chiang, Wei-Lin and Wu, Tianhao and Gonzalez, Joseph E. and Kadous, M. Waleed and Stoica, Ion},
  booktitle = {International Conference on Learning Representations},
  year = {2025},
  url = {https://openreview.net/forum?id=8sSqNntaMr}
}

@inproceedings{li2023selectivecontext,
  title = {Compressing Context to Enhance Inference Efficiency of Large Language Models},
  author = {Li, Yucheng and Dong, Bo and Guerin, Frank and Lin, Chenghua},
  booktitle = {Proceedings of the 2023 Conference on Empirical Methods in Natural Language Processing},
  pages = {6342--6353},
  year = {2023},
  address = {Singapore},
  publisher = {Association for Computational Linguistics},
  doi = {10.18653/v1/2023.emnlp-main.391},
  url = {https://aclanthology.org/2023.emnlp-main.391/}
}

@inproceedings{yuan2026videostar,
  title = {Video-{STAR}: Reinforcing Open-Vocabulary Action Recognition with Tools},
  author = {Yuan, Zhenlong and Qu, Xiangyan and Qian, Chengxuan and Chen, Rui and Tang, Jing and Sun, Lei and Chu, Xiangxiang and Zhang, Dapeng and Wang, Yiwei and Cai, Yujun and Li, Shuo},
  booktitle = {The Fourteenth International Conference on Learning Representations},
  year = {2026},
  url = {https://openreview.net/forum?id=NBOHB6aYZh}
}

@inproceedings{yuan2026autodriver,
  title = {AutoDrive-{R}$^2$: Incentivizing Reasoning and Self-Reflection Capacity for {VLA} Model in Autonomous Driving},
  author = {Yuan, Zhenlong and Qian, Chengxuan and Tang, Jing and Chen, Rui and Song, Zijian and Sun, Lei and Chu, Xiangxiang and Cai, Yujun and Zhang, Dapeng and Li, Shuo},
  booktitle = {The Fourteenth International Conference on Learning Representations},
  year = {2026},
  url = {https://openreview.net/forum?id=KVWaCzJrrq}
}

@inproceedings{yuan2025dvpmvs,
  title = {{DVP-MVS}: Synergize Depth-Edge and Visibility Prior for Multi-View Stereo},
  author = {Yuan, Zhenlong and Luo, Jinguo and Shen, Fei and Li, Zhaoxin and Liu, Cong and Mao, Tianlu and Wang, Zhaoqi},
  booktitle = {Proceedings of the AAAI Conference on Artificial Intelligence},
  volume = {39},
  number = {9},
  pages = {9743--9752},
  year = {2025},
  doi = {10.1609/aaai.v39i9.33056},
  url = {https://doi.org/10.1609/aaai.v39i9.33056}
}

@article{yuan2024tsarmvs,
  title = {{TSAR-MVS}: Textureless-aware Segmentation and Correlative Refinement Guided Multi-View Stereo},
  author = {Yuan, Zhenlong and Cao, Jiakai and Wang, Zhaoqi and Li, Zhaoxin},
  journal = {Pattern Recognition},
  volume = {154},
  pages = {110565},
  year = {2024},
  doi = {10.1016/j.patcog.2024.110565},
  url = {https://doi.org/10.1016/j.patcog.2024.110565}
}

@inproceedings{chen2024bimcvr,
  title = {{BIMCV-R}: A Landmark Dataset for 3D {CT} Text-Image Retrieval},
  author = {Chen, Yinda and Liu, Che and Liu, Xiaoyu and Arcucci, Rossella and Xiong, Zhiwei},
  booktitle = {Medical Image Computing and Computer Assisted Intervention -- MICCAI 2024},
  pages = {124--134},
  publisher = {Springer},
  year = {2024},
  doi = {10.1007/978-3-031-72120-5_12},
  url = {https://doi.org/10.1007/978-3-031-72120-5_12}
}

@inproceedings{chen2023selfsupervisedneuron,
  title = {Self-Supervised Neuron Segmentation with Multi-Agent Reinforcement Learning},
  author = {Chen, Yinda and Huang, Wei and Zhou, Shenglong and Chen, Qi and Xiong, Zhiwei},
  booktitle = {Proceedings of the Thirty-Second International Joint Conference on Artificial Intelligence},
  pages = {609--617},
  publisher = {International Joint Conferences on Artificial Intelligence Organization},
  year = {2023},
  doi = {10.24963/ijcai.2023/68},
  url = {https://www.ijcai.org/proceedings/2023/68}
}

\appendix

\section{Task Splits and Evaluation Protocol}
\label{app:task-splits}

\begin{figure*}[t]
    \includegraphics[width=1\linewidth]{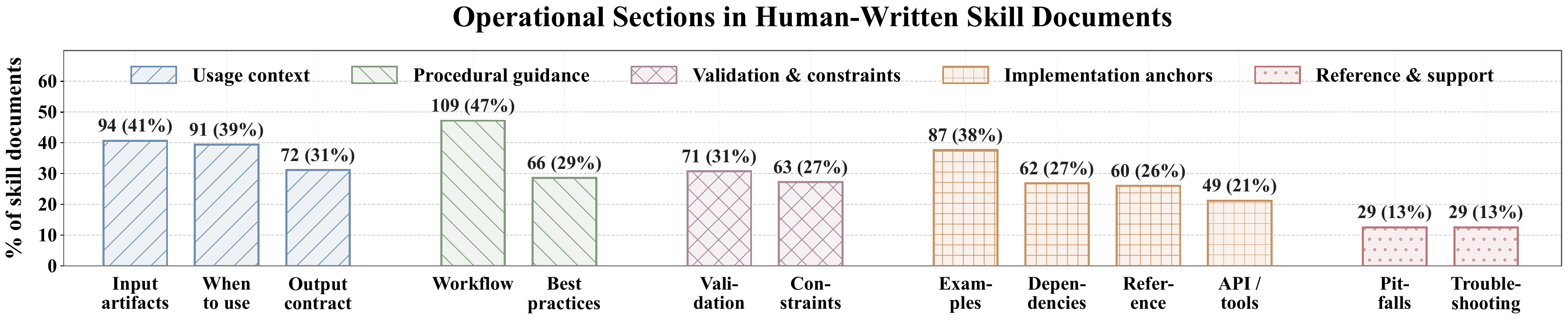}
    \caption{
    Distribution of operational section types in human-written SkillsBench skills.
Bars show the percentage of 231 \texttt{SKILL.md} files containing each normalized section type, with raw counts annotated.
Section types are grouped into usage context, procedural guidance, validation and constraints, implementation anchors, and reference/support material.
The breadth of these sections motivates task-conditioned preservation rather than uniform compression.
    }
    \label{fig:skill-section-distribution}
\end{figure*}

We separate corpus-level skill profiling from performance-based model selection and final evaluation.
The full SkillsBench corpus contains 88 tasks with human-written skills. We use all 88 tasks only for unlabeled structural profiling, such as counting skill files, section types, token lengths, code density, validation cues, and constraint cues. No verifier score or agent outcome from the held-out evaluation split is used during this profiling step.

For execution-based experiments, we use the 86 tasks whose environments and verifiers are runnable under our evaluation harness. These tasks are partitioned into three roles. First, 28 tasks are used for template calibration, where we estimate the behavior of fixed rewrite strategies and remove clearly unstable templates. Second, 38 tasks are used for policy learning and adaptation, where the strategy selector is fit from task/skill features and economic utility. Third, 20 tasks are reserved as the main held-out evaluation panel. The learned policy is frozen before being applied to this panel.

Cross-agent robustness experiments use the 86 runnable tasks, but only after the rewrite policy has been fixed. These runs therefore test transfer across agent/model stacks rather than providing additional data for policy fitting. We report them separately from the 20-task held-out comparison.

We report the held-out panel in Table~\ref{tab:appendix-heldout-panel}. 
The columns describe only information available before evaluation:
\begin{itemize}
    \item \textbf{\#Skills}: Number of skill files provided with the task.
    \item \textbf{Skill Tok.}: Total token count of the original human-written skills.
    \item \textbf{Code}: Fraction of skill tokens appearing in code blocks.
    \item \textbf{Val.}: Whether the skill set contains explicit validation or checking guidance.
    \item \textbf{Constr.}: Whether the skill set contains explicit constraint or rule guidance.
    \item \textbf{Policy selection signal}: Structural signal used by the frozen policy to select a rewrite behavior.
\end{itemize}
These signals are computed from task and skill features, not from held-out execution outcomes.
\begin{table*}[h]
\centering
\footnotesize
\setlength{\tabcolsep}{2.1pt}
\renewcommand{\arraystretch}{1.03}

\newcolumntype{Y}{>{\raggedright\arraybackslash}X}
\newcolumntype{L}[1]{>{\raggedright\arraybackslash}p{#1}}
\newcolumntype{C}[1]{>{\centering\arraybackslash}p{#1}}
\newcolumntype{R}[1]{>{\raggedleft\arraybackslash}p{#1}}

\begin{tabularx}{\textwidth}{@{}
L{0.19\textwidth}
L{0.15\textwidth}
R{0.052\textwidth}
R{0.060\textwidth}
R{0.050\textwidth}
C{0.043\textwidth}
C{0.055\textwidth}
Y
@{}}
\toprule
\textbf{Task} &
\textbf{Domain} &
\makecell[r]{\textbf{\#}\\\textbf{Skills}} &
\textbf{Skill Tok.} &
\textbf{Code} &
\textbf{Val.} &
\textbf{Rule} &
\textbf{Policy selection signal} \\
\midrule

\seqsplit{adaptive-cruise-control}
& Control / optimization
& 5 & 1681 & 0.606 & Y & N
& Preserve implementation anchors and simulation checks \\

\seqsplit{civ6-adjacency-optimizer}
& Control / optimization
& 4 & 3246 & 0.331 & Y & Y
& Preserve API calls, grid rules, and validation cues \\

\seqsplit{earthquake-plate-calculation}
& Scientific computing
& 1 & 1115 & 0.504 & N & Y
& Preserve formulas, units, and threshold rules \\

\seqsplit{econ-detrending-correlation}
& Data analysis / ML
& 1 & 788 & 0.248 & N & N
& Preserve statistical transformation logic \\

\seqsplit{energy-ac-optimal-power-flow}
& Control / optimization
& 3 & 2564 & 0.396 & Y & Y
& Preserve solver/API anchors and feasibility checks \\

\seqsplit{energy-market-pricing}
& Control / optimization
& 4 & 3034 & 0.538 & N & Y
& Preserve market equations and implementation conventions \\

\seqsplit{fix-erlang-ssh-cve}
& Code / debugging / build
& 6 & 8415 & 0.550 & Y & Y
& Preserve code references, commands, and patch constraints \\

\seqsplit{flood-risk-analysis}
& Data analysis / ML
& 3 & 1766 & 0.311 & N & Y
& Preserve risk formulas and decision thresholds \\

\seqsplit{glm-lake-mendota}
& Scientific computing
& 3 & 1491 & 0.382 & N & N
& Preserve domain equations and parameter conventions \\

\seqsplit{grid-dispatch-operator}
& Control / optimization
& 3 & 2206 & 0.558 & N & Y
& Preserve dispatch APIs and operational constraints \\

\seqsplit{invoice-fraud-detection}
& Spreadsheet / office
& 3 & 3585 & 0.419 & Y & Y
& Preserve spreadsheet operations and audit checks \\

\seqsplit{lab-unit-harmonization}
& Data analysis / ML
& 1 & 1805 & 0.252 & N & N
& Preserve normalization rules and unit mappings \\

\seqsplit{mars-clouds-clustering}
& Scientific computing
& 3 & 1460 & 0.621 & N & N
& Preserve scientific feature definitions and formulas \\

\seqsplit{pptx-reference-formatting}
& Spreadsheet / office
& 1 & 4473 & 0.126 & N & N
& Preserve file-format operations and layout anchors \\

\seqsplit{r2r-mpc-control}
& Control / optimization
& 4 & 872 & 0.369 & N & N
& Preserve controller implementation anchors \\

\seqsplit{react-performance-debugging}
& Web / visualization
& 2 & 1836 & 0.165 & N & Y
& Preserve debugging workflow and constraint cues \\

\seqsplit{sales-pivot-analysis}
& Spreadsheet / office
& 2 & 2299 & 0.565 & N & Y
& Preserve spreadsheet APIs and aggregation constraints \\

\seqsplit{simpo-code-reproduction}
& Data analysis / ML
& 2 & 1454 & 0.517 & N & Y
& Preserve algorithmic objective and formula constraints \\

\seqsplit{threejs-structure-parser}
& Web / visualization
& 2 & 1085 & 0.462 & N & Y
& Preserve parser APIs and output constraints \\

\seqsplit{travel-planning}
& Data analysis / ML
& 6 & 476 & 0.212 & N & N
& Preserve compact decision rules over workflow detail \\

\bottomrule
\end{tabularx}

\caption{
Main held-out evaluation panel.
All structural features are computed from the original human-written skills before rewriting.
The policy selection signal summarizes the pre-execution evidence used by the frozen rewrite policy; it is not chosen from held-out outcomes.
}
\label{tab:appendix-heldout-panel}
\end{table*}
\section{Full Economic Metrics}
\label{app:economic-metrics}

This appendix gives the complete set of economic diagnostics used in our analysis. 
For a task $\tau$ and skill condition $a$, let $q_\tau^a$ denote the verifier partial score, $b_\tau^a \in \{0,1\}$ denote the binary reward when available, $C_{\mathrm{skill},\tau}^a$ denote direct skill-token cost, $C_{\mathrm{agent},\tau}^a$ denote downstream agent-token cost, and $C_\tau^a=C_{\mathrm{skill},\tau}^a+C_{\mathrm{agent},\tau}^a$ denote total token cost. 
The original human-written skill condition is denoted by $a=0$.
All ratios are computed relative to the original-skill baseline for the same task and agent stack before macro-averaging across tasks.

\paragraph{Layer 1: quality.}
We report both verifier partial score $q_\tau^a$ and binary reward $b_\tau^a$ when the benchmark exposes a full-success signal. 
The main quality-normalized quantity is quality retention,
$\mathrm{QR}_\tau^a=q_\tau^a/\max(q_\tau^0,\epsilon)$,
which allows us to distinguish harmless cost reductions from rewrites that save tokens by degrading task performance. 
We also use the one-sided quality loss $\mathrm{QL}_\tau^a=\max(0,1-\mathrm{QR}_\tau^a)$ as a diagnostic in ablations.

\paragraph{Layer 2: token accounting.}
The direct skill-token ratio $r_{s,\tau}^a$ and downstream agent-token ratio $r_{a,\tau}^a$ decompose total cost into the prompt-side cost of providing skills and the execution-side cost incurred after the agent begins acting. 
The total cost ratio $\rho_\tau^a=C_\tau^a/\max(C_\tau^0,\epsilon)$ is the primary cost metric. 
This decomposition is important because a rewrite can reduce $C_{\mathrm{skill}}$ while increasing $C_{\mathrm{agent}}$, which means apparent prompt savings may not translate into a cheaper agent run.

\paragraph{Layer 3: economic diagnostics.}
Beyond the metrics reported in the main text, we use the following diagnostics:
\begin{align}
\mathrm{SCD}_\tau^a &= 1-r_{s,\tau}^a, \\
\mathrm{DI}_\tau^a &= r_{a,\tau}^a-r_{s,\tau}^a, \\
\mathrm{TTR}_\tau^a &= \frac{r_{a,\tau}^a}{\max(r_{s,\tau}^a,\epsilon)}, \\
\mathrm{QCF}_\tau^a &= \mathrm{QR}_\tau^a-\rho_\tau^a, \\
\mathrm{BPT}_\tau^a &= \frac{\mathrm{QR}_\tau^a}{\max(\rho_\tau^a,\epsilon)}, \\
\mathrm{GRE}_\tau^a &= 
\mathbb{I}[\mathrm{QR}_\tau^a \ge 1-\delta]\,
\frac{b_\tau^a}{\max(\rho_\tau^a,\epsilon)} .
\end{align}
Here, $\mathrm{SCD}$ is the skill-compression dividend: positive values indicate that the rewritten skills are shorter than the original skills. 
$\mathrm{DI}$ is downstream inflation: positive values indicate that downstream execution cost falls less than direct skill cost, or increases despite shorter skills. 
$\mathrm{TTR}$ is the token-transfer ratio; values above $1$ indicate that prompt-side compression is partially transferred into execution-side cost. 
$\mathrm{QCF}$ is a quality--cost frontier score, positive only when quality retention exceeds total cost ratio. 
$\mathrm{BPT}$ measures quality retained per unit total cost. 
$\mathrm{GRE}$ is a stricter reward-based economy score, used only when quality is near-lossless.

\begin{table*}[t]
\centering
\small
\setlength{\tabcolsep}{4.2pt}
\renewcommand{\arraystretch}{1.03}
\begin{tabular}{lrrrrrrrrrrr}
\toprule
\textbf{Condition} &
\textbf{Partial} &
\textbf{QR} &
\boldmath$\rho$ &
\boldmath$r_s/r_a$ &
\textbf{SCD} &
\textbf{DI} &
\textbf{EO} &
\textbf{TTR} &
\textbf{QCF} &
\textbf{BPT} &
\textbf{NLD} \\
\midrule
Original skills
& \second{0.815} & \second{1.00} & 1.00 & 1.00 / 1.00 & 0.00 & \best{0.00} & 0.00 & \best{1.00} & 0.00 & 1.00 & 0.00 \\
API/code anchored
& 0.792 & 0.97 & 0.94 & \second{0.60} / \second{0.96} & \second{0.40} & 0.36 & 0.00 & 1.60 & \second{0.03} & \second{1.03} & \second{0.04} \\
Rule/formula anchored
& 0.555 & 0.68 & \best{0.92} & 0.65 / \best{0.94} & 0.35 & \second{0.29} & 0.00 & \second{1.45} & -0.24 & 0.74 & 0.00 \\
Workflow-guarded
& 0.731 & 0.90 & 1.11 & \best{0.58} / 1.14 & \best{0.42} & 0.56 & 0.14 & 1.97 & -0.21 & 0.81 & 0.00 \\
Policy-selected
& \best{0.819} & \best{1.01} & \second{0.93} & 0.62 / \best{0.94} & 0.38 & 0.32 & 0.00 & 1.52 & \best{0.08} & \best{1.09} & \best{0.07} \\
\bottomrule
\end{tabular}
\caption{
Full economic diagnostics on the primary held-out evaluation stack.
Partial is mean verifier partial score.
QR is quality retention.
$\rho$ is total cost ratio.
$r_s/r_a$ reports direct skill-token and downstream agent-token ratios.
SCD is skill-compression dividend, DI is downstream inflation, EO is execution overrun, TTR is token-transfer ratio, QCF is quality--cost frontier score, BPT is bang-per-token, and NLD is near-lossless dividend.
Nonlinear quantities are computed per task and then macro-averaged.
}
\label{tab:full-economic-metrics}
\end{table*}

Table~\ref{tab:full-economic-metrics} illustrates why skill rewriting cannot be evaluated by prompt length alone. 
All rewritten conditions reduce direct skill tokens, but their downstream behavior differs sharply. 
Workflow-guarded rewrites have the largest skill-compression dividend, yet they also have the highest downstream inflation and positive execution overrun, leading to a worse total cost ratio. 
Rule/formula anchoring is economical but loses too much quality on implementation-sensitive tasks. 
API/code anchoring gives a stronger quality--cost balance, but still transfers part of the saved prompt cost into downstream execution. 
The policy-selected condition is not the shortest condition; instead, it gives the best frontier score and bang-per-token by choosing which operational anchors to preserve for each task.

\section{Held-out Quality-Preserving Cost Savings}
\label{app:cost-saving}

\begin{figure*}[t]
\centering
\includegraphics[width=1\linewidth]{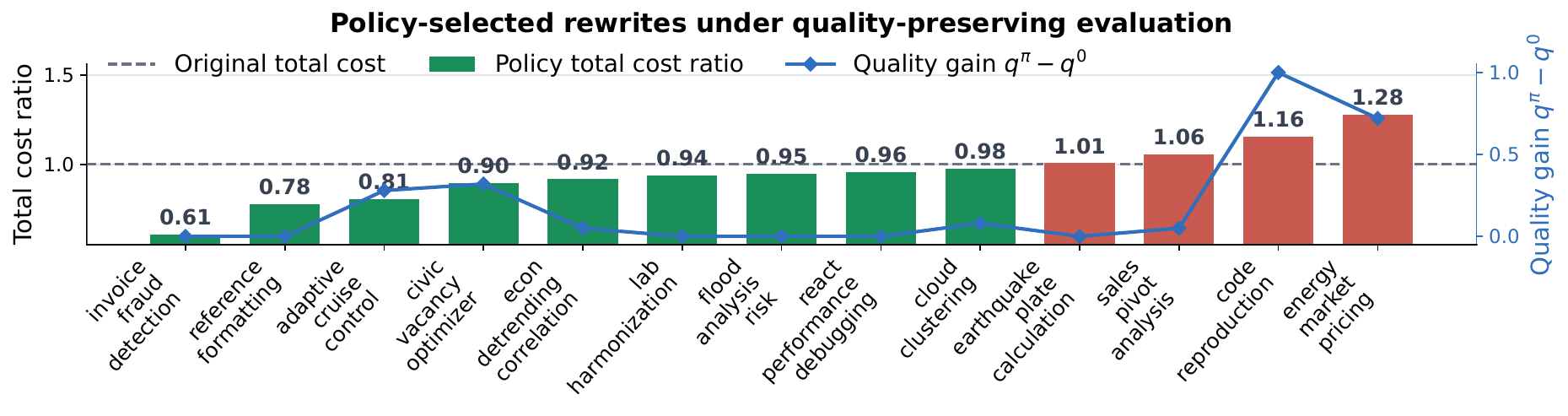}
\caption{
Quality-preserving cost savings of policy-selected rewrites.
We include scored held-out tasks where the policy-selected rewrite does not reduce verifier partial score relative to the original-skill baseline.
Bars show total cost ratio; values below $1$ indicate total token savings.
The blue line shows quality gain $q^{\pi}-q^0$.
}
\label{fig:quality-preserving-cost}
\end{figure*}

We report full held-out results and failure classifications for the clean main evaluation. 
Each task is evaluated under five skill conditions: original human-written skills, API/code anchoring, rule/formula anchoring, workflow guarding, and the frozen policy-selected rewrite. 
Table~\ref{tab:full-heldout-summary} reports the aggregate token counts for the primary 20-task held-out Gemini Flash evaluation. These aggregate counts are a held-out-only diagnostic and should be distinguished from the four-setting average reported in the abstract. On this held-out panel, policy-selected rewriting reduces direct skill tokens from 45.7K to 28.2K, downstream API tokens from 720.1K to 680.2K, and total tokens from 765.8K to 708.4K. This corresponds to a 38.3\% direct skill-token reduction, a 5.5\% downstream-token reduction, and a 7.5\% aggregate total-token reduction while slightly improving mean partial score.
The policy-selected condition obtains the highest mean partial score while reducing aggregate total tokens relative to the original-skill baseline. Fixed templates show less balanced behavior: rule/formula anchoring has low aggregate cost but much lower mean partial score, while workflow guarding increases downstream API usage despite shorter skills.
\begin{table*}[t]
\centering
\small
\setlength{\tabcolsep}{5pt}
\renewcommand{\arraystretch}{1.03}
\begin{tabular}{lrrrrrr}
\toprule
\textbf{Skill condition} &
\textbf{Mean partial} &
\textbf{Skill tok.} &
\textbf{API tok.} &
\textbf{Total tok.} &
\boldmath$r_a$ &
\boldmath$\rho_{\mathrm{agg}}$ \\
\midrule
Original skills
& \second{0.8147} & 45{,}651 & 720{,}139 & 765{,}790 & 1.00 & 1.00 \\
API/code anchored
& 0.7921 & \second{27{,}277} & 694{,}601 & 721{,}878 & \second{0.96} & 0.94 \\
Rule/formula anchored
& 0.5546 & 29{,}632 & \best{675{,}122} & \best{704{,}754} & \best{0.94} & \best{0.92} \\
Workflow-guarded
& 0.7307 & \best{26{,}688} & 824{,}274 & 850{,}962 & 1.14 & 1.11 \\
Policy-selected
& \best{0.8188} & 28{,}160 & \second{680{,}213} & \second{708{,}373} & \best{0.94} & \second{0.93} \\
\bottomrule
\end{tabular}
\caption{
Held-out-only aggregate token summary on the 20-task Gemini Flash panel.
$\rho_{\mathrm{agg}}$ is computed from aggregate total tokens in this table and is therefore a held-out diagnostic; the headline 12.8\% total-cost reduction is computed by averaging the four policy-selected settings in Table~\ref{tab:main-results}.
}
\label{tab:full-heldout-summary}
\end{table*}
Figure~\ref{fig:quality-preserving-cost} focuses on the quality-preserving regime. 
We include scored held-out tasks for which the policy-selected rewrite does not reduce verifier partial score relative to the original skills. 
Bars show total cost ratio, where values below $1$ indicate lower total token cost than the original-skill baseline. 
The blue line shows the quality gain $q^{\pi}-q^0$. This diagnostic is computed per task and is not used for headline aggregate results.
Most tasks in this subset fall below the baseline cost line, showing that the policy often converts skill rewriting into genuine total-cost savings rather than merely shorter prompts. 
The few above-baseline cases are informative: they indicate settings where quality is preserved or improved, but downstream execution still becomes more expensive. 
This is consistent with our cost-transfer argument and motivates reporting total execution cost rather than skill length alone.

\section{Qualitative Case Studies}
\label{app:case-studies}

We provide representative case studies comparing original human-written skills with policy-selected rewrites. 
The goal of these examples is not to claim that individual cases prove the aggregate result, but to show the mechanism behind the quantitative trends: effective rewrites preserve task-relevant operational anchors, while removing background exposition that does not help execution. 
For each case, we report verifier outcomes and token costs from the clean main evaluation. 
The skill excerpts are shortened for readability; ellipses indicate omitted lines from the same skill file.

\begin{table*}[h]
\centering
\footnotesize
\setlength{\tabcolsep}{2.2pt}
\renewcommand{\arraystretch}{1.04}

\newcolumntype{L}[1]{>{\raggedright\arraybackslash}p{#1}}
\newcolumntype{C}[1]{>{\centering\arraybackslash}p{#1}}
\newcolumntype{R}[1]{>{\raggedleft\arraybackslash}p{#1}}

\begin{tabularx}{\textwidth}{@{}
L{0.18\textwidth}
X
R{0.060\textwidth}
R{0.065\textwidth}
R{0.070\textwidth}
R{0.065\textwidth}
R{0.070\textwidth}
C{0.120\textwidth}
@{}}
\toprule
\textbf{Case} &
\textbf{Main effect} &
\textbf{Base $q$} &
\textbf{Policy $q$} &
\textbf{Skill ratio} &
\textbf{API ratio} &
\textbf{Total ratio} &
\textbf{Tests} \\
\midrule

\seqsplit{adaptive-cruise-control}
& Quality + cost win
& 0.667 & 1.000 & 0.963 & 0.805 & 0.812
& 8/12 $\rightarrow$ 12/12 \\

\seqsplit{civ6-adjacency-optimizer}
& Quality + moderate cost win
& 0.900 & 1.000 & 0.672 & 0.930 & 0.918
& 9/10 $\rightarrow$ 10/10 \\

\seqsplit{invoice-fraud-detection}
& Same quality, large cost saving
& 1.000 & 1.000 & 0.526 & 0.621 & 0.613
& 2/2 $\rightarrow$ 2/2 \\

\seqsplit{pptx-reference-formatting}
& Same quality, large skill compression
& 1.000 & 1.000 & 0.107 & 0.825 & 0.780
& 12/12 $\rightarrow$ 12/12 \\

\seqsplit{sales-pivot-analysis}
& Cost-transfer diagnostic
& 0.522 & 0.565 & 0.732 & 1.109 & 1.076
& 12/23 $\rightarrow$ 13/23 \\

\seqsplit{r2r-mpc-control}
& Failure diagnostic
& 1.000 & 0.667 & 0.944 & 1.119 & 1.114
& 6/6 $\rightarrow$ 4/6 \\

\bottomrule
\end{tabularx}

\caption{
Qualitative case summary.
$q$ is verifier partial score. Skill/API/total ratios are relative to the original-skill baseline for the same task.
Diagnostic cases are included to show cost transfer and policy failures rather than only favorable examples.
}
\label{tab:case-summary}
\end{table*}

\begin{table*}[h]
\centering
\small
\setlength{\tabcolsep}{4pt}
\renewcommand{\arraystretch}{1.03}
\begin{tabular}{p{0.22\linewidth}p{0.34\linewidth}p{0.36\linewidth}}
\toprule
\textbf{Task} & \textbf{Policy-preserved anchors} & \textbf{Observed behavior} \\
\midrule
adaptive-cruise-control
& PID interface, control law, clamping, stability checks
& Quality improves to full pass while API tokens drop, indicating fewer failed tuning/recovery steps. \\

civ6-adjacency-optimizer
& Pruning rules, tile scoring, anchor search, constraint validation
& The rewrite turns a broad search heuristic into an executable optimization plan. \\

invoice-fraud-detection
& Spreadsheet APIs, fuzzy matching workflow, formula/error validation
& Broad spreadsheet guidance is removed, but audit-critical operations remain; quality is unchanged and cost drops sharply. \\

pptx-reference-formatting
& OOXML tools, slide XML structure, formatting anchors, thumbnail validation
& Very large skill compression is possible because the task needs a narrow subset of the original PPTX manual. \\

sales-pivot-analysis
& Pivot-table API anchors, cacheId rule, field mapping
& Partial score improves slightly, but downstream API usage rises, showing cost transfer. \\

r2r-mpc-control
& LQR recursion and MPC workflow, but limited derivation detail
& The rewrite is too compact for a mathematically dense control task; quality and cost both worsen. \\
\bottomrule
\end{tabular}
\caption{
Mechanistic interpretation of case-study outcomes.
}
\label{tab:case-study-mechanisms}
\end{table*}

\paragraph{Quality and cost improvements.}
The first two cases show that policy-selected rewrites can improve both task quality and total cost. 
In \textit{adaptive-cruise-control}, the original skills contain the necessary PID and vehicle-control knowledge, but the relevant control interface, clamping behavior, and validation targets are distributed across explanatory text. 
The policy rewrite makes these operational anchors explicit, yielding a full verifier pass and reducing total cost to 0.812 of the baseline. 
In \textit{civ6-adjacency-optimizer}, the policy rewrite preserves the same high-level three-stage search strategy but exposes the executable pieces: candidate pruning, scoring, anchor search, and final constraint validation. 
This changes the outcome from one failed test to a full pass while reducing total cost by 8.2\%.

\paragraph{Near-lossless compression.}
The office-document cases illustrate that large prompt reductions are possible when the rewrite preserves the task-relevant subset of a broad skill manual. 
For \textit{invoice-fraud-detection}, the original spreadsheet skill includes extensive financial-modeling conventions, while the task mainly requires extraction, matching, spreadsheet I/O, and validation. 
The policy rewrite keeps spreadsheet APIs and audit-relevant checks, preserving full quality while reducing total cost by 38.7\%. 
For \textit{pptx-reference-formatting}, the original PPTX skill is a general manual for creation, editing, design, and XML manipulation. 
The rewrite keeps only the OOXML tools, file-structure anchors, formatting operations, and thumbnail validation needed for the reference-formatting task, reducing skill tokens to 10.7\% of the original while retaining full verifier quality.

\paragraph{Cost transfer and failure cases.}
The final two cases explain why our evaluation uses total cost rather than skill length alone. 
In \textit{sales-pivot-analysis}, the policy rewrite keeps the crucial pivot-table anchors, including \texttt{cacheId=0}, cache definitions, and field mapping. 
This improves partial score from 0.522 to 0.565, but downstream API tokens increase by 10.9\%, producing a total cost ratio of 1.076. 
This is a clean cost-transfer case: the rewritten skill is shorter, but the agent spends more execution tokens resolving the task. 
In \textit{r2r-mpc-control}, the policy rewrite preserves the Riccati recursion and MPC loop, but compresses mathematical context that appears necessary for stable control design. 
Quality drops from 1.000 to 0.667 and total cost rises to 1.114 of baseline. 
This failure supports our central framing: skill rewriting should be learned as cost-aware operational knowledge preservation, not as generic summarization.

\section{Agent Stacks, Runtime, and Cost}
\label{app:agent-stacks}

We evaluate the frozen rewrite policy across four agent/model stacks. 
For runtime and provider-cost accounting, we summarize the original and policy-selected conditions on the 86 runnable SkillsBench tasks. Fixed-strategy results are reported in Table~\ref{tab:main-results}, while this appendix table focuses on raw token volume, runtime, and monetary cost.

All rewritten skill variants are generated once using Gemini-3-Flash-Preview with the controlled rewrite prompts described in Appendix~\ref{app:rewrite-prompts}. The generated skill files are then frozen and reused unchanged across all agent stacks. This design separates the \emph{rewrite model} from the \emph{execution model}: differences in downstream quality and cost come from how each agent uses the same skill condition, not from regenerating skills separately for each model.

\paragraph{Agent stacks.}
Table~\ref{tab:agent-stack-details} summarizes the four evaluation stacks. 
The stacks cover different agent regimes: high-throughput tool use, stronger Gemini-family reasoning, code-oriented file editing, and long-context coding. 
This design follows the SkillsBench finding that performance and cost vary substantially across commercial agent harnesses, even when the task and skill condition are held fixed.

\begin{table*}[h]
\centering
\small
\setlength{\tabcolsep}{4pt}
\renewcommand{\arraystretch}{1.03}
\begin{tabular}{p{0.25\linewidth}p{0.21\linewidth}p{0.24\linewidth}p{0.21\linewidth}}
\toprule
\textbf{Agent stack} & \textbf{Evaluation role} & \textbf{Behavioral profile} & \textbf{Cost profile} \\
\midrule
Gemini CLI + Gemini 3 Flash Preview
& Primary evaluation stack
& Fast tool-use agent with high throughput; tends to use more exploratory execution.
& Low unit token cost, higher token volume, short wall-clock runtime. \\

Gemini CLI + Gemini 3 Pro
& Gemini-family robustness stack
& Stronger reasoning than Flash; requires fewer exploration rounds on many tasks.
& Higher unit token cost, substantially lower token volume than Flash. \\

Codex + GPT-5.4
& Code-oriented robustness stack
& Strong file editing and implementation behavior; particularly suited to API/code-heavy tasks.
& Higher per-run cost, but often fewer implementation retries. \\

Claude Code + Opus 4.6
& Long-context robustness stack
& Strong coding and tool-use ability with more deliberate execution.
& High runtime variance; strong quality but higher timeout sensitivity. \\
\bottomrule
\end{tabular}
\caption{
Agent/model stacks used for cross-stack evaluation.
The same frozen policy-selected rewrites are evaluated under all stacks.
}
\label{tab:agent-stack-details}
\end{table*}

\paragraph{Reproducibility fields.}
We report model identifiers, agent CLI versions, decoding settings, timeout and retry policy, tool permissions, evaluation dates, and price tables because commercial agent stacks may change over time.

\paragraph{Token and runtime accounting.}
For each trajectory, we record direct skill tokens, downstream API input/output tokens, API request count, wall-clock runtime, verifier partial score, and binary reward when available. 
Total token usage is computed as the sum of direct skill tokens and downstream API tokens. 

Monetary cost is computed from the provider price table used at evaluation time; because provider prices change, we report normalized cost ratios in the main analysis and retain dollar costs only as reproducibility metadata.

A key point is that token volume is not monotonic in model strength. 
In the original SkillsBench evaluation, Gemini 3 Flash consumes substantially more input tokens than Gemini 3 Pro despite being the cheaper model. 
This reflects a common agentic pattern: smaller or faster models compensate with more iterative exploration, while stronger models often solve with fewer rounds. 
Our cross-stack results preserve this distinction, so lower token usage for Gemini 3 Pro than Gemini 3 Flash is expected rather than anomalous.

\begin{table*}[h]
\centering
\small
\setlength{\tabcolsep}{3.5pt}
\renewcommand{\arraystretch}{1.03}
\begin{tabular}{llrrrrr}
\toprule
\textbf{Stack} &
\textbf{Condition} &
\makecell{\textbf{Input tok.}\\\textbf{/ run}} &
\makecell{\textbf{Output tok.}\\\textbf{/ run}} &
\makecell{\textbf{Total tok.}\\\textbf{/ run}} &
\makecell{\textbf{Runtime}\\\textbf{/ run}} &
\makecell{\textbf{Monetary}\\\textbf{cost ratio}} \\
\midrule
Gemini 3 Flash
& Original  & \second{1.075M} & \second{12.1K} & \second{1.087M} & \second{7.5 min} & \second{1.00}  \\
& Policy & \best{0.972M} & \best{11.8K} & \best{0.984M} & \best{6.8 min} & \best{0.91}  \\
\midrule
Gemini 3 Pro
& Original & \second{0.465M} & \second{10.9K} & \second{0.476M} & \second{8.9 min} & \second{1.00}  \\
& Policy  & \best{0.426M} & \best{10.6K} & \best{0.437M} & \best{8.1 min} & \best{0.92}  \\
\midrule
Codex + GPT-5.4
& Original & \second{1.087M} & \second{11.6K} & \second{1.099M} & \second{10.8 min} & \second{1.00}  \\
& Policy & \best{0.967M} & \best{11.3K} & \best{0.978M} & \best{9.6 min} & \best{0.89}  \\
\midrule
Claude Code + Opus 4.6
& Original  & \second{1.448M} & -- & -- & \second{14.5 min} & \second{1.00}  \\
& Policy  & \best{1.265M} & -- & -- & \best{13.1 min} & \best{0.87}  \\
\bottomrule
\end{tabular}
\caption{
Runtime and cost summary across agent/model stacks on the 86-task runnable panel.
For each task and condition, token counts and runtime are first averaged over valid trials and then macro-averaged across tasks.
The monetary cost ratio is normalized to the original-skill condition within the same stack and is computed using provider-side input/output token accounting.
Claude Code output-token counts are omitted because session logs do not reliably recover completion-token counts; its monetary cost ratio is computed from logged provider-side accounting.
}
\label{tab:cross-stack-cost}
\end{table*}

\begin{table}[h]
\centering
\footnotesize
\setlength{\tabcolsep}{3pt}
\renewcommand{\arraystretch}{1.03}
\begin{tabular}{p{0.22\columnwidth}p{0.70\columnwidth}}
\toprule
\textbf{Stage} & \textbf{Condensed log evidence} \\
\midrule
Run metadata
& Codex + GPT-5.4 on \texttt{sales-pivot-analysis}; wall-clock time 8.23 minutes, including 6.19 minutes of agent execution. \\

Skill routing
& The agent loads the spreadsheet skill and identifies pivot-table construction requirements, including the cache and worksheet constraints needed for valid Excel pivots. \\

Document routing
& The agent loads the PDF skill and routes table extraction through \texttt{pdfplumber}, instead of treating the PDF as unstructured text. \\

Data inspection
& The trajectory inspects \texttt{income.xlsx}, extracts population tables from \texttt{population.pdf}, and checks the schema before generating the workbook. \\

Artifact construction
& The agent creates \texttt{demographic\_analysis.xlsx} with \texttt{SourceData} and four analysis sheets: population, earners, regions, and income quartiles. \\

Anchor execution
& Spreadsheet anchors are converted into executable workbook operations: each analysis sheet receives a pivot-table definition rather than a static summary table only. \\

Self-verification
& The workbook is reopened successfully; each analysis sheet contains one pivot definition, with 2,454 source rows and balanced quartile assignments. \\

Data caveat
& The log records 49 rows with missing income-derived fields, corresponding to blank income rows and unmatched territory entries. \\

Cost record
& The run consumes 334,630 input tokens, 158,336 cached input tokens, and 13,507 output tokens, with a logged cost of \$0.683. \\
\bottomrule
\end{tabular}
\caption{
Condensed execution log for a representative Codex + GPT-5.4 trajectory.
The log illustrates how skill-level anchors become concrete tool choices, validation checks, and artifact-level operations during agent execution.
}
\label{tab:codex-gpt54-log-example}
\end{table}

\paragraph{Cross-stack cost behavior.}
The results show that policy-selected rewrites reduce total cost across all four stacks while preserving or improving mean verifier partial score. 
The reduction is largest for code-oriented stacks, where removing irrelevant skill text while preserving API and implementation anchors decreases both prompt-side cost and downstream retries. 
Gemini 3 Flash remains the most token-intensive Gemini stack, consuming more than twice the input tokens of Gemini 3 Pro, but its lower unit token cost makes it competitive in monetary cost. 
Gemini 3 Pro uses fewer tokens because it performs less exploratory tool use, while Codex and Opus exhibit higher absolute runtime due to more complex file-editing and verification behavior.

These results support the main conclusion of the paper: the economic value of skill rewriting is not captured by skill-token compression alone. 
A rewrite is useful only when it preserves the operational anchors that reduce downstream search, debugging, and recovery. 
Across model families, policy-selected rewrites consistently reduce direct skill cost and usually reduce downstream execution cost as well, showing that the learned policy transfers beyond the primary training stack.

\section{Policy Learning Details}
\label{app:policy-details}

This appendix provides implementation details for the task-conditioned rewrite policy introduced in Section~\ref{sec:method}. The policy is designed to answer a practical question: given a new task and its original skill set, which information-preservation strategy should be used before running the agent? We intentionally use a low-capacity, interpretable selector rather than a black-box reinforcement-learning system, because the available feedback is expensive, sparse, and heterogeneous across tasks.

\subsection{Learning Signal}

Each policy example is a completed task--strategy run. For a task $\tau$ and rewrite strategy $a$, we observe verifier quality, binary reward when available, direct skill-token cost, downstream agent-token cost, and execution validity. Runs that fail due to infrastructure or harness errors are removed from the learning set; runs with valid verifier outcomes are retained even when task performance is poor, since such failures are informative about the strategy.

The scalar learning target is the economic utility used in the main method:

\begin{equation}
\label{eq:app-policy-utility}
\begin{aligned}
U_\tau^a
&=
\mathrm{QR}_{\tau}^{a}
+\lambda_{\mathrm{save}}(1-\rho_{\tau}^{a}) \\
&\quad
-\lambda_{\mathrm{over}}\mathrm{EO}_{\tau}^{a}
+\lambda_{\mathrm{nld}}\mathrm{NLD}_{\tau}^{a}.
\end{aligned}
\end{equation}

Here $\mathrm{QR}$ is quality retention, $\rho$ is total cost ratio, $\mathrm{NLD}$ is the near-lossless dividend, and $\mathrm{EO}$ is downstream execution overrun. 
We set
$\lambda_{\mathrm{save}}=1.0$,
$\lambda_{\mathrm{over}}=1.0$,
$\lambda_{\mathrm{nld}}=0.5$, and
$\delta=0.05$.
These values are selected using only the 28 template-calibration tasks. After selection, they are fixed for policy adaptation, held-out evaluation, and cross-model transfer. The 38 policy-adaptation tasks are used only to fit the strategy selector under this fixed utility definition; they are not used to retune utility weights or the near-lossless tolerance. Held-out and cross-model runs are used only for evaluation.

This target makes the policy sensitive to the main economic failure mode in our setting: a rewrite may shorten the skill file while increasing downstream exploration.

\subsection{Policy Inputs}

For each task, the selector receives a static profile $\phi_\tau$ computed before any agent execution. The profile combines skill-structure features, domain features, and risk indicators. Table~\ref{tab:policy-features} summarizes the feature groups.

\begin{table}[h]
\centering
\small
\setlength{\tabcolsep}{3pt}
\renewcommand{\arraystretch}{1.03}
\begin{tabular}{p{0.28\columnwidth}p{0.62\columnwidth}}
\toprule
\textbf{Feature group} & \textbf{Examples} \\
\midrule
Skill size
& Number of skills, total skill tokens, median skill length, long-skill indicator. \\

Implementation anchors
& Code-token ratio, API/tool mentions, commands, imports, code blocks, library-specific object names. \\

Procedural anchors
& Ordered steps, validation checks, constraints, pitfalls, recovery instructions, file-format requirements. \\

Rule anchors
& Formulas, thresholds, schemas, units, tie-breaks, invariants, scientific conventions. \\

Task family
& Data analysis, scientific computing, software debugging, control/optimization, spreadsheet/office, web/visual interaction. \\

Execution risk
& Docker or dependency risk, multimodal input, long-context requirement, verifier strictness proxy. \\
\bottomrule
\end{tabular}
\caption{
Feature groups used by the policy selector.
All features are computed from the task instruction and original skill files before running the agent.
}
\label{tab:policy-features}
\end{table}

These features are not intended to describe the task exhaustively. Instead, they expose the structural cues that determine which information is economically valuable to preserve. For example, code-heavy skills often require API/code anchoring, while scientific and data-analysis tasks often depend on formulas, schemas, units, and exact conventions.
Such anchor sensitivity also appears beyond text-only agents. 
Multimodal and embodied tasks rely on geometry, visibility, motion, and trajectory cues for multi-view stereo, video reasoning, and autonomous driving \citep{yuan2024tsarmvs,yuan2025dvpmvs,yuan2026videostar,yuan2026autodriver}, while biomedical tasks depend on modality-specific conventions, supervision signals, and evaluation protocols \citep{chen2023selfsupervisedneuron,chen2024bimcvr}. 
This motivates tracking domain and anchor features separately from raw token count.


\subsection{Candidate Actions}

The action space consists of rewrite strategies that survived calibration. Source-native compacting is retained as a diagnostic baseline, but it is not used as a final policy arm because early calibration showed that preserving the original organization without targeted anchor protection was unstable. The final selector chooses among:
\begin{itemize}
    \item \textbf{API/code anchoring}, which preserves implementation details such as imports, object construction, API calls, commands, and minimal executable snippets;
    \item \textbf{Rule/formula anchoring}, which preserves equations, schemas, thresholds, units, constraints, and domain conventions;
    \item \textbf{Workflow guarding}, which preserves ordered procedures, validation checks, failure modes, and recovery cues.
\end{itemize}

This action space reflects the central hypothesis of the paper: rewrite strategies should differ in what they preserve, not merely in how much text they remove.

\subsection{Learning Procedure}

We cast strategy selection as an offline contextual bandit. The learner receives tuples $(\phi_\tau,a,U_\tau^a)$ from calibration and adaptation runs, where $\phi_\tau$ is the pre-execution task profile, $a$ is a rewrite strategy, and $U_\tau^a$ is the observed economic utility. The final selector is trained in three stages.

First, we aggregate repeated runs for the same task--strategy pair using mean utility and retain auxiliary statistics such as valid rate, median cost ratio, and quality retention. Second, we fit an action-conditioned utility model over normalized numeric features and Boolean indicators. Third, we distill the fitted preferences into a small ordered decision list used as the frozen deployed policy. A rule is accepted only if it improves mean utility on the adaptation set and has sufficient support. This distillation step is important: it prevents the policy from memorizing individual tasks and makes the learned strategy auditable.

In Algorithm~\ref{alg:policy-learning}, $\mathcal{D}_{\mathrm{train}}$ contains runs from the template-calibration and policy-adaptation splits after utility hyperparameters have been fixed.

\begin{algorithm}[t]
\small
\caption{Cost-aware policy learning}
\label{alg:policy-learning}
\begin{algorithmic}[1]
\Require Policy-training runs $\mathcal{D}_{\mathrm{train}}$ from template-calibration and policy-adaptation splits, task profiles $\phi_\tau$, strategies $\mathcal{A}_{\pi}$
\State Remove infrastructure failures and runs without valid verifier accounting

\For{each task $\tau$ and strategy $a$}
    \State Compute $q_\tau^a$, $\rho_\tau^a$, $\mathrm{QR}_\tau^a$, $\mathrm{NLD}_\tau^a$, and $\mathrm{EO}_\tau^a$
    \State Compute utility $U_\tau^a$ using Eq.~\ref{eq:app-policy-utility}
\EndFor

\State Fit an action-conditioned utility model $\widehat{U}(\phi,a)$
\State Initialize the default action as the strategy with highest mean utility
\State Greedily add feature rules that improve held-in utility with minimum support
\State Freeze the resulting decision list before held-out evaluation
\State \Return Frozen decision-list policy distilled from $\widehat{U}(\phi,a)$
\end{algorithmic}
\end{algorithm}
\subsection{Final Selector}

The frozen selector used in held-out and cross-model evaluation is intentionally compact. It is distilled from the learned utility model into a small decision list, so that its behavior can be inspected directly. The default strategy is API/code anchoring, reflecting its strong general performance on heterogeneous tool-use and implementation-heavy tasks. The selector overrides this default when calibration and adaptation feedback indicate that another preservation family is more economically reliable: rule/formula anchoring for scientific or schema-heavy tasks, and workflow guarding for validation-heavy procedural tasks.

\begin{table}[h]
\centering
\small
\setlength{\tabcolsep}{3pt}
\renewcommand{\arraystretch}{1.05}
\begin{tabular}{p{0.36\columnwidth}p{0.25\columnwidth}p{0.27\columnwidth}}
\toprule
\textbf{Condition} & \textbf{Selected strategy} & \textbf{Rationale} \\
\midrule
Default
& API/code anchoring
& Preserve executable anchors for heterogeneous tool-use tasks. \\

Scientific / control or formula-heavy
& Rule/formula anchoring
& Preserve equations, units, thresholds, and domain conventions. \\

Schema or statistical-definition-heavy data analysis
& Rule/formula anchoring
& Preserve schemas, aggregation rules, statistical definitions, and invariants. \\

Validation-heavy procedural workflow
& Workflow guarding
& Preserve ordered checks, recovery cues, and verifier-facing validation steps. \\
\bottomrule
\end{tabular}
\caption{
Frozen selector used for held-out and cross-model evaluation.
The selector is deliberately low-capacity: it chooses a preservation family before rewriting, rather than producing task-specific rewrite rules.
}
\label{tab:frozen-policy}
\end{table}

Although the final decision list is short, it is distilled from multi-strategy economic feedback rather than chosen in hindsight. Its simplicity is part of the experimental design: the goal is to test whether quality--cost signals identify stable preservation preferences that transfer beyond the calibration runs. The selector therefore operates as a lightweight layer above the rewrite strategies, choosing which class of operational anchors to preserve before any agent execution occurs.
\subsection{Leakage Control}

The policy uses only task metadata and original skill files as inputs. It does not access verifier outputs, generated solutions, agent traces, or downstream token usage for the held-out task at inference time. Economic outcomes are used only during calibration to learn the selector. After the decision list is frozen, held-out tasks are rewritten once according to $\pi(\phi_\tau)$ and then evaluated with the same task instructions, environments, and verifiers as the original-skill baseline. This protocol ensures that the policy is adaptive to task and skill structure, but not adaptive to held-out verifier results.

\section{Rewrite Strategies and Prompts}
\label{app:rewrite-prompts}

This appendix describes the LLM-assisted rewriting protocol. Each prompt specifies a preservation objective, section structure, protected anchors, and audits; the task instruction, environment, and verifier are never rewritten.
\subsection{Strategy Families}

We design rewrite strategies as information-preservation families. Each strategy asks the rewriter to make a different class of operational anchors salient while removing low-value prose. Table~\ref{tab:rewrite-prompt-families} summarizes the strategy families used in calibration and evaluation.

\begin{table}[h]
\centering
\small
\setlength{\tabcolsep}{3pt}
\renewcommand{\arraystretch}{1.03}
\begin{tabular}{p{0.25\columnwidth}p{0.29\columnwidth}p{0.34\columnwidth}}
\toprule
\textbf{Strategy} & \textbf{Preservation objective} & \textbf{Typical sections} \\
\midrule
Source-native compacting
& Preserve the author's original organization while removing redundancy.
& Original outline, operational details, validation and risks. \\

Workflow guarding
& Make procedures, checks, constraints, and failure modes explicit.
& When to use, inputs/outputs, core workflow, validation checks, pitfalls. \\

API/code anchoring
& Preserve executable API, library, command, and object-construction patterns.
& Required artifacts, API/tool patterns, implementation anchors, workflow, validation. \\

Rule/formula anchoring
& Preserve exact definitions, formulas, schemas, thresholds, units, and invariants.
& Exact definitions, decision rules, formulas, minimal implementation anchors, failure modes. \\
\bottomrule
\end{tabular}
\caption{
Rewrite strategy families.
Each strategy defines what information should remain salient after rewriting, rather than imposing a universal surface template.
}
\label{tab:rewrite-prompt-families}
\end{table}

Source-native compacting is used as a diagnostic strategy: it tests whether preserving the original skill organization alone is sufficient. The main policy arms focus on targeted anchor preservation, because calibration showed that generic compaction can shorten skills without protecting the information that prevents downstream exploration.

\subsection{Canonical Rewrite Prompt}

All rewrite prompts follow the same high-level schema. The only strategy-specific fields are the strategy identifier, section plan, preservation policy, target compression range, and anchor budget. The prompt requests a structured JSON response so that the rewritten skill and audit metadata can be stored separately.



\begin{tcolorbox}[
    title=Canonical Rewrite Prompt Skeleton,
    fonttitle=\bfseries,
    colback=gray!5,
    colframe=black!70,
    boxrule=0.5pt,
    left=6pt, right=6pt, top=4pt, bottom=4pt
]

\footnotesize

Rewrite the original \texttt{SKILL.md} using the specified strategy family.

\medskip
\textbf{Inputs:}
\begin{itemize}[nosep, leftmargin=1.2em]
    \item Strategy id: \texttt{\{strategy\_id\}}
    \item Target sections: \texttt{\{section\_list\}}
    \item Preservation policy: \texttt{\{strategy\_policy\}}
\end{itemize}

\medskip
\textbf{Return} one JSON object with fields:
\texttt{rewritten\_skill\_md},
\texttt{preserved\_information},
\texttt{compressed\_information},
and \texttt{risk\_flags}.

\medskip
\textbf{Rules:}
\begin{itemize}[nosep, leftmargin=1.2em]
    \item Preserve YAML metadata.
    \item Do not solve the task; do not invent facts.
    \item Preserve exact: file types, API names, commands, class/function names, constants, formulas, validation thresholds, and ``do not'' rules.
    \item Keep minimal executable code/API patterns when APIs are non-obvious.
    \item Keep coordinate systems, schemas, formulas, units, invariants, and validation checks for scientific or rule-governed skills.
    \item Use concise Markdown suitable for \texttt{SKILL.md}.
\end{itemize}

\end{tcolorbox}

This prompt structure is intentionally conservative. The rewriter is asked to transform only the skill document, not to infer a new solution procedure for the task. The explicit JSON fields also make failures inspectable: if a rewrite compresses important information, the missing anchor usually appears in the risk flags or audit checks.

\subsection{Strategy-Specific Prompt Policies}

Table~\ref{tab:strategy-prompt-policies} gives the strategy-specific policies used to instantiate the canonical prompt. The compression targets are soft targets rather than hard truncation limits. This matters because overly rigid length control can remove exactly the anchors that make a skill useful.

\begin{table}[h]
\centering
\footnotesize
\setlength{\tabcolsep}{3pt}
\renewcommand{\arraystretch}{1.03}
\begin{tabular}{p{0.25\columnwidth}p{0.16\columnwidth}p{0.47\columnwidth}}
\toprule
\textbf{Strategy} & \textbf{Target ratio} & \textbf{Prompt policy} \\
\midrule
Source-native compacting
& 0.72
& Preserve the source author's information architecture; merge sections only when redundant; compress within existing sections. \\

Workflow guarding
& 0.62
& Make the procedure easy to follow; keep validation, constraints, recovery cues, and pitfalls visible. \\

API/code anchoring
& 0.82
& Preserve complete minimal API and code-construction patterns when APIs are non-obvious; remove only low-value examples and prose. \\

Rule/formula anchoring
& 0.78
& Preserve formulas, thresholds, schemas, units, ranking rules, tie-breaks, invariants, and compact examples defining correctness. \\
\bottomrule
\end{tabular}
\caption{
Strategy-specific prompt policies.
The ratios are soft compression targets; the primary constraint is preservation of the strategy's protected anchors.
}
\label{tab:strategy-prompt-policies}
\end{table}

The API/code strategy is allowed a higher token ratio because implementation anchors are often long but high-value. Conversely, workflow guarding uses a lower target ratio because procedural prose can often be compressed if checks and pitfalls remain explicit. Rule/formula anchoring sits between these extremes: formulas and schemas must be exact, but surrounding explanation can usually be shortened.

\subsection{Post-Generation Audits and Repair}

After generation, each rewritten skill is checked by lightweight audits before it is materialized as a benchmark variant. These audits are not used to score the final task; they are used only to ensure that the rewrite respects the preservation protocol.

\begin{table}[h]
\centering
\small
\setlength{\tabcolsep}{3pt}
\renewcommand{\arraystretch}{1.03}
\begin{tabular}{p{0.28\columnwidth}p{0.58\columnwidth}}
\toprule
\textbf{Audit} & \textbf{Purpose} \\
\midrule
Token-ratio check
& Records the rewritten-to-original skill-token ratio and flags large expansions or unexpectedly aggressive compression. \\

Code/API term coverage
& Measures whether important code-like terms, commands, constants, imports, and API identifiers remain present. \\

Code-block retention
& Checks whether source code blocks or their leading executable patterns are preserved when required. \\

Anchor restoration
& If a protected code/API block is missing, a compact source-derived anchor block is appended to the rewritten skill. \\

Short-skill guard
& If an already compact original skill becomes longer without clear benefit, the original skill is retained. \\

Text hygiene
& Normalizes malformed punctuation and encoding artifacts before writing the final \texttt{SKILL.md}. \\
\bottomrule
\end{tabular}
\caption{
Post-generation audits applied to rewritten skills.
The audits enforce preservation constraints without using held-out verifier feedback.
}
\label{tab:rewrite-audits}
\end{table}

The most important repair step is anchor restoration. When the source skill contains high-value executable material that is absent from the rewrite, the pipeline appends a compact ``Implementation Anchors'' block derived from the original skill. This makes the rewrite procedure robust to a common failure mode of LLM summarization: fluent compression that omits the exact API pattern needed for execution.

\subsection{Materializing Benchmark Variants}

For each task and strategy, the pipeline copies the original task directory and rewrites only files under \texttt{environment/skills/}. All other task components are unchanged. The resulting directory is therefore a controlled task variant: any performance or cost difference between variants is attributable to the skill condition rather than changes in task instructions, data files, environment setup, or verifier logic.

Formally, for task $\tau=(x,e,v,\mathcal{S}_\tau)$ and strategy $a$, the materialized variant is
\begin{equation}
\widetilde{\tau}^{a}
=
(x,e,v,R_a(\mathcal{S}_\tau)).
\end{equation}
This controlled materialization is what allows our evaluation to interpret downstream token changes as an economic effect of skill rewriting. A rewrite may reduce direct skill tokens, but if it removes anchors that the agent needs, the agent may spend more tokens recovering those details during execution.

\subsection{Relation to the Learned Policy}

The rewrite strategies are the action space for the policy in Appendix~\ref{app:policy-details}. The policy does not generate new prompt text at inference time. Instead, it selects one of the pre-defined preservation objectives based on task and skill features, and the corresponding rewrite prompt is applied. This separation keeps the system interpretable: the learned component decides \emph{which kind of information to preserve}, while the rewrite prompt specifies \emph{how that preservation objective is enforced}.

\begin{table*}[h]
\centering
\footnotesize
\setlength{\tabcolsep}{3.2pt}
\renewcommand{\arraystretch}{1.04}
\begin{tabular}{p{0.15\linewidth}p{0.39\linewidth}p{0.39\linewidth}}
\toprule
\textbf{Task} & \textbf{Original skill excerpt} & \textbf{Policy-selected rewrite excerpt} \\
\midrule

adaptive-cruise-control
&
\textbf{PID Controller Implementation}

Overview: A PID controller is a feedback control mechanism...

Control Law:
\texttt{output = Kp * error + Ki * integral(error) + Kd * derivative(error)}

Discrete-Time Implementation:
\texttt{class PIDController: ... compute(error, dt) ...}

Anti-Windup: Integral windup occurs when output saturates...
&
\textbf{Capability and use cases}

Feedback control for adaptive cruise control and vehicle speed regulation.

\textbf{API/tool patterns}

\texttt{class PIDController: ... compute(error, dt)}

\textbf{Implementation anchors}

Control law; error definition; output clamping.

\textbf{Validation checks}

Steady-state error, overshoot, stability, \texttt{dt > 0}. \\

\midrule

civ6-adjacency-optimizer
&
\textbf{Map-Based Constraint Optimization Strategy}

Why exhaustive search fails: combinatorial explosion...

The Three-Phase Strategy:
Phase 1 prune the search space.
Phase 2 identify high-value spots.
Phase 3 anchor point search.

Algorithm Skeleton:
\texttt{def optimize\_placements(...)}
&
\textbf{Capability and use cases}

Placement optimization on spatial grids where exhaustive search is intractable.

\textbf{Required artifacts}

\texttt{map\_tiles}, hard constraints, number of placements.

\textbf{Implementation anchors}

Pruning criteria: invalid, dominated, isolated.
Scoring: intrinsic value + adjacency potential + cluster potential.
Anchor logic for center-constrained problems.

\textbf{Validation checks}

Terrain, range limits, blocked tiles, mutual exclusion. \\

\midrule

invoice-fraud-detection
&
\textbf{Requirements for Outputs}

Zero formula errors; preserve existing templates.

\textbf{Financial models}

Color coding standards, number formatting standards, formula construction rules.

\textbf{XLSX creation, editing, and analysis}

Reading and analyzing data; pandas; openpyxl; formulas; formatting...
&
\textbf{Capability and use cases}

Create, edit, and analyze spreadsheet files; financial modeling; formula validation.

\textbf{Required artifacts}

\texttt{pandas}, \texttt{openpyxl}, \texttt{recalc.py}.

\textbf{API/tool patterns}

Read with pandas; write formulas with openpyxl; run recalculation.

\textbf{Implementation anchors}

Color conventions, number formats, no hardcoded formulas, source comments.

\textbf{Workflow}

Load/create, modify, save, recalculate, verify. \\

\bottomrule

pptx-reference-formatting
&
\textbf{PPTX creation, editing, and analysis}

Overview: a PPTX is a ZIP archive containing XML files.

Reading and analyzing content:
text extraction with \texttt{markitdown}; raw XML access; unpacking.

Creating a new presentation:
design principles, color palette selection, layout guidance...
&
\textbf{Capability and use cases}

Creation, editing, and analysis of PPTX files.

\textbf{Required artifacts}

\texttt{unpack.py}, \texttt{pack.py}, \texttt{validate.py}, \texttt{rearrange.py}, \texttt{inventory.py}, \texttt{replace.py}, \texttt{thumbnail.py}.

\textbf{API/tool patterns}

Extract text; unpack/edit/validate/pack OOXML; generate thumbnails.

\textbf{Implementation anchors}

OOXML structure, web-safe fonts, layout constraints.

\textbf{Validation checks}

Thumbnail check for cutoff, overlap, positioning, contrast. \\

\end{tabular}
\caption{
Original and policy-selected skill excerpts for six case studies, Part 1 of 2.
This table shows the first four case studies. The policy rewrites reorganize skills around executable anchors, including required artifacts, API/tool patterns, implementation details, workflow steps, validation checks, and pitfalls.
}
\label{tab:case-skill-excerpts-part1}
\end{table*}

\begin{table*}[h]
\centering
\footnotesize
\setlength{\tabcolsep}{3.2pt}
\renewcommand{\arraystretch}{1.04}
\begin{tabular}{p{0.15\linewidth}p{0.39\linewidth}p{0.39\linewidth}}
\toprule
\textbf{Task} & \textbf{Original skill excerpt} & \textbf{Policy-selected rewrite excerpt} \\
\midrule

sales-pivot-analysis
&
\textbf{XLSX Creation, Editing, and Analysis}

Read data with pandas or openpyxl.

Create Excel files with formatting.

\textbf{Creating Pivot Tables}

Critical note: all pivot tables must use \texttt{cacheId=0}.

Basic pivot table structure with cache, fields, table definition...
&
\textbf{API/tool patterns}

Analysis pattern; cell-level pattern; creation pattern.

\textbf{Implementation anchors}

Pivot table construction:
\texttt{CacheDefinition}, \texttt{CacheField}, \texttt{TableDefinition}, \texttt{Location}, \texttt{PivotField}, \texttt{DataField}.

\textbf{Critical rule}

Use \texttt{cacheId=0}; field indices must match cache fields.

\textbf{Workflow}

Prepare source data; build cache; define table; map fields; verify workbook. \\

\midrule

r2r-mpc-control
&
\textbf{Finite-Horizon LQR for MPC}

Problem formulation:
\texttt{J = sum x'Qx + u'Ru + x\_N' P x\_N}

Backward Riccati recursion:
\texttt{K\_k = inv(R + B'PB) B'PA}

Forward simulation:
\texttt{u\_k = -K\_k x\_k}

Python implementation and MPC application...
&
\textbf{Capability and use cases}

Minimize quadratic cost over horizon N for MPC control.

\textbf{API/tool patterns}

A compact \path|finite_horizon_lqr(A,B,Q,R,N,x0)| stub.

\textbf{Implementation anchors}

Backward Riccati recursion; forward simulation equations.

\textbf{Validation checks}

Initialize terminal cost \texttt{P\_N}.

\textbf{Pitfalls}

Improper terminal initialization can cause instability. \\

\bottomrule
\end{tabular}
\caption{
Original and policy-selected skill excerpts for six case studies, Part 2 of 2.
This table continues Table~\ref{tab:case-skill-excerpts-part1} with the remaining two case studies. The r2r-mpc-control example illustrates a limitation: preserving formulas without sufficient derivation and stability context may be insufficient for mathematically dense control tasks.
}
\label{tab:case-skill-excerpts-part2}
\end{table*}









\end{document}